\documentclass[a4paper,fleqn]{cas-sc}

\usepackage[numbers]{natbib}

\usepackage[table]{xcolor} 
\usepackage{amsmath} 
\usepackage{graphicx}
\usepackage{subfigure}
\usepackage{adjustbox}
\usepackage{placeins}
\usepackage{multirow}
\usepackage{booktabs}
\usepackage{threeparttable}
\usepackage{float}
\usepackage{wrapfig}
\usepackage[toc]{appendix}
\usepackage{booktabs}
\usepackage{tabularx}
\usepackage{array}
\usepackage{algorithm}
\usepackage{algpseudocode}
\usepackage{caption}
\usepackage{graphicx}
\usepackage[left]{lineno}
\def\tsc#1{\csdef{#1}{\textsc{\lowercase{#1}}\xspace}}
\tsc{WGM}
\tsc{QE}

\begin{document}








\let\WriteBookmarks\relax
\def\floatpagepagefraction{1}
\def\textpagefraction{.001}

\title[mode=title]{Curvature-aware dynamic precision approach for physics-informed neural networks}
\shorttitle{}
\shortauthors{}

\author[1,2]{Yingjie Shao}[orcid=0009-0008-2083-0497]\ead{yingjie.shao@wur.nl}
\author[2]{Ioannis N. Athanasiadis}[orcid=0000-0003-2764-0078]
\author[1]{George van Voorn}[orcid=0000-0002-1369-3964]
\author[2]{Taniya Kapoor\corref{cor1}}[orcid=0000-0002-6361-446X]\ead{taniya.kapoor@wur.nl}

\affiliation[1]{organization={Mathematical \& Statistical Methods Group (Biometris), Wageningen University \& Research},
                city={Wageningen},
                postcode={6700AA},
                country={The Netherlands}}

\affiliation[2]{organization={Artificial Intelligence Group, Wageningen University \& Research},
                city={Wageningen},
                postcode={6700AA},
                country={The Netherlands}}

\cortext[cor1]{Corresponding author}

\begin{abstract} Physics-informed neural networks (PINNs) have become a promising framework for simulating partial differential equations (PDEs) by embedding physical laws directly into neural network training. However, recent studies show that PINN optimisation is sensitive to numerical precision. Existing implementations commonly use either single precision (FP32), which is computationally efficient but prone to failure modes, or double precision (FP64), which is robust but substantially expensive. This creates a trade-off between computational efficiency and numerical accuracy. To reduce the computational cost of double-precision training while retaining prediction accuracy, we propose a curvature-aware precision controller that adapts numerical precision during training rather than treating it as a fixed implementation choice. The proposed method reuses curvature information derived from the limited-memory BFGS (L-BFGS) optimiser to construct a precision controller, retaining FP32 when lower precision is sufficient and promoting computation to FP64 when the training dynamics indicate numerical sensitivity or precision-limited stagnation. We evaluate the proposed approach on four canonical PINN failure-mode benchmarks and an irradiance-driven ordinary differential equation example. We further test the proposed approach across different neural network architectures. The method consistently matches or even slightly exceeds full FP64 solution accuracy while reducing training time relative to full double-precision training on all benchmark equations. The obtained results indicate that precision sensitivity in PINN optimisation is phase-dependent, and that selectively applying higher precision only during numerically critical stages can lower computational cost without sacrificing predictive accuracy. \\
\end{abstract}

\begin{highlights}
\item We propose a dynamic precision approach for training physics-informed neural networks and showcase its applicability on benchmark failure-mode equations.
\item We reuse L-BFGS curvature information to build a precision-switching controller.
\item The proposed dynamic approach preserves double precision, FP64-level, accuracy at lower cost.
\item The proposed controller is architecture-agnostic.
\end{highlights}


\maketitle
\section{Introduction}
Partial differential equations (PDEs) are fundamental in many scientific disciplines and model several spatio-temporal phenomena, including fluid dynamics, heat transfer, wave propagation, ecological dynamics, and biological pattern formation~\citep{abbasPDEModelsVegetation2025,burzawaAccelerationPDEBasedBiological2020,mattheij2005partial}, among others. Physics-informed neural networks (PINNs) have recently emerged as a promising framework for simulating PDEs by incorporating governing equations, initial conditions, and boundary conditions directly into the training loss~\citep{RAISSI2019686}. For the forward-problem setting of PINN that will be addressed in this study, the governing equations and associated conditions are assumed to be known, and PINNs are employed to approximate the unknown solutions by enforcing the PDE residuals at collocation points~\citep{doi:10.1137/19M1274067,RAISSI2019686}. This mesh-free methodology offers an alternative to traditional grid-based numerical solvers and enables solution approximations to be evaluated at arbitrary spatio-temporal locations after training~\citep{cuomoScientificMachineLearning2022, khanra2026physics}.

Despite their flexibility, jointly optimising multiple loss components during PINN training remains challenging. The PDE residual and boundary and initial condition terms are embedded in a single loss function, creating a complex multi-objective optimisation landscape in which different loss components have different scales and convergence rates, leading to ill-conditioned optimisation dynamics~\citep{rathore2024challenges,wang2022and, xiang2022self}. As a result, the gradient of the composite loss for training may be dominated by some constraints while others remain poorly optimised, making optimiser choice important for obtaining an accurate PDE solution~\citep{jnini2026curvature,rathore2024challenges}. This difficulty is further increased by the need to compute derivatives of the neural-network output through automatic differentiation and backpropagation, especially for PDEs involving higher-order differential operators~\citep{kapoor2023physics, datar2026fast}. Consequently, PINNs may exhibit failure modes in which the training loss, especially the PDE residual loss, appears to converge while the predicted solution remains far from the true PDE solution, particularly for complex systems and high-dimensional parameter spaces~\citep{xu2025fp64,krishnapriyan2021characterizing}. Since these failure modes are closely linked to the optimisation difficulty, the choice of optimiser becomes an important factor in PINN training~\citep{rathore2024challenges,krishnapriyan2021characterizing}.

\begin{figure}
    \centering
    \includegraphics[width=0.99\linewidth]{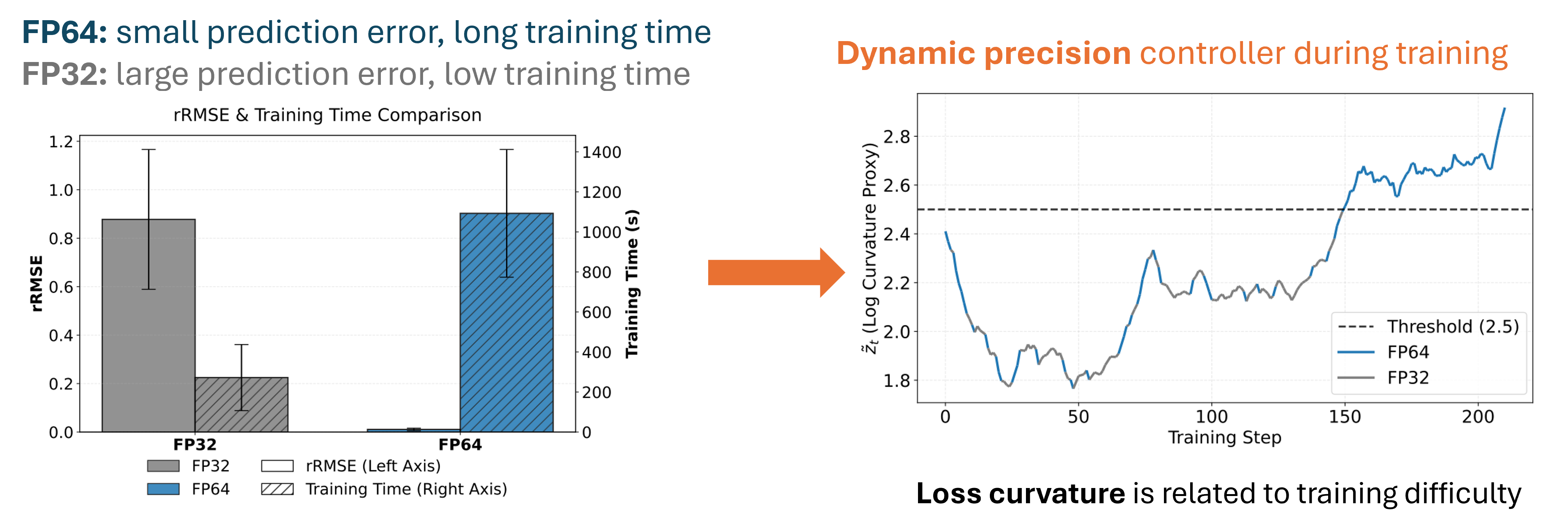}
    \captionsetup{hypcap=false}
    \captionof{figure}{(Left) Trade-off between prediction accuracy and training speed when using different training precisions. (Right) As loss curvature has been widely used to characterise training difficulty and ill-conditioning in PINNs, we propose a curvature-aware controller that adjusts numerical precision during training. The controller switches to FP64 (the blue part in the curve) in difficult training phases (high curvature) based on curvature signals derived from the L-BFGS optimiser.}\label{fig:motivation}
\end{figure}

Among the optimisers commonly used for PINN training, L-BFGS is particularly sensitive to numerical precision because it uses curvature information constructed from a limited history of parameter and gradient updates~\citep{Nocedal1980UpdatingQM,liuLimitedMemoryBFGS1989,rathore2024challenges,jnini2026curvature}. In optimisation, loss curvature reflects how rapidly the loss landscape changes with respect to parameter updates and is commonly associated with ill-conditioning and optimisation difficulty. When function or gradient evaluations are inaccurate, classical BFGS and L-BFGS updates can be corrupted, and line-search behaviour can become unreliable~\citep{doi:10.1137/20M1373190,doi:10.1137/19M1240794}. Recent studies suggest that some PINN failure modes are related to the premature termination of L-BFGS optimisation due to insufficient numerical precision~\citep{xu2025fp64}. This observation is consistent with the broader scientific-computing view that lower-precision arithmetic can improve performance but must be used carefully when numerical accuracy is important~\citep{kashiMixedprecisionNumericsScientific2026}. In PINNs, this numerical sensitivity is further amplified by derivative-based PDE residuals, which require derivatives of the neural-network output to be computed through automatic differentiation~\citep{RAISSI2019686,cuomoScientificMachineLearning2022}. Double precision (FP64) provides a more stable numerical regime and has been shown to mitigate precision-related failures in PINN optimisation~\citep{xu2025fp64}.

However, full FP64 training is computationally costly, especially on modern hardware that is optimised for lower-precision arithmetic, resulting in a substantial performance gap between FP64 and lower-precision formats~\citep{micikevicius2018mixed,kashiMixedprecisionNumericsScientific2026}. Existing approaches partially address the broader challenge of improving PINN training efficiency and robustness through mixed-precision training~\citep{xue2022novel}, adaptive collocation point selection~\citep{pinn_importancesampling,tang2023pinns,WU2023115671}, or improved model architectures~\citep{xu2025sub}. Many adaptive sampling and collocation methods focus on heterogeneous difficulty across the spatial or spatio-temporal domain by reallocating residual points toward regions where the PDE residual or solution error is larger~\citep{pinn_importancesampling,tang2023pinns,WU2023115671}. For example, \citet{tang2023pinns} dynamically adjusts the collocation density through adaptive sampling, while \citet{wang2022and} adaptively updates the weights of different loss components during training. Recent work further suggests that optimisation difficulty and numerical sensitivity can also vary across different phases of training dynamics, since a model may escape a failure mode simply by switching to higher precision when FP32 training stagnates~\citep{xu2025fp64}.

To balance the trade-off between computational cost and prediction accuracy, we investigate whether PINNs can achieve accuracy comparable to full FP64 training while reducing computational cost through adaptive precision control (The motivation of this project is summarised in Figure~\ref{fig:motivation}). In this work, we propose reusing curvature information from the history of the second-order optimiser L-BFGS~\citep{Nocedal1980UpdatingQM,liuLimitedMemoryBFGS1989} as an online signal for dynamically switching precision during training. Unlike prior work, which mostly focuses on layer-wise or static precision allocation, our method adapts precision based on optimisation dynamics. We investigate whether numerical precision can be linked to the training phase and adjusted based on curvature information derived from the L-BFGS history. The main contributions of this article are summarised as follows:
\begin{itemize}
\item We propose an approach that balances the trade-off between computational cost and prediction accuracy for PINN training.
\item Our approach uses a curvature-aware precision controller that reuses information from the L-BFGS history as an online signal for precision control in PINN training.
\item We introduce an architecture-agnostic approach that recovers much of the robustness of FP64 training at a lower wall-clock cost for several state-of-the-art physics-informed neural network-based methods.
\end{itemize}

The rest of the article is structured as follows. We first describe previous work related to mixed precision and how curvature information is used in neural network optimisation, as well as its link to the ill-conditioning problem in PINNs, in Section~\ref{sec:related work}. The general PINN setup is then defined in Section~\ref{sec:PINN}, followed by the proposed curvature-aware dynamic precision approach in Section~\ref{sec:curvature_aware}. The approach is evaluated across different model structures described in Section~\ref{sec:Network_architectures_and_evaluation}, and specific experiment setups and results are discussed in Section~\ref{sec:result_discussion}, followed by the conclusion presented in Section~\ref{sec:conclusion}.

\section{Related work}\label{sec:related work}

Mixed-precision training: Mixed-precision methods are widely used in deep learning and scientific computing to improve throughput and reduce memory cost while retaining sufficient numerical accuracy~\citep{micikevicius2018mixed,kashiMixedprecisionNumericsScientific2026}. Advanced physics-informed learning strategies based on transformer or Mamba-based architectures face memory challenges~\citep{gao2025mlpinn}. Some recent work has explored adaptive precision strategies, such as layer-wise or operator-wise precision allocation~\citep{kummer2023adaptive,rajagopal2020multi}, but these approaches are not designed to explicitly adapt precision to different phases of training dynamics, explicitly leverage optimisation geometry for precision scheduling, or target PINN-specific structures. Some recent methods have incorporated second-order information into precision allocation. For example, HAWQ and its follow-up work use the Hessian spectrum or Hessian-trace-based sensitivity metrics with automatic bit selection to determine relative layer-wise quantisation precision in deep networks~\citep{dong2019hawq,NEURIPS2020_d77c7035}. This suggests that curvature-related signals can provide useful guidance for precision switching.

Although automatic mixed precision (AMP) is widely used in deep learning, it primarily relies on FP16 computation with loss scaling to mitigate underflow. Prior work shows that vanilla AMP can be unstable for PINNs, where higher-order derivative terms amplify numerical sensitivity and require derivative-aware scaling beyond standard loss scaling~\citep{xue2022novel,GLADSTONE2025106161}. Furthermore, standard mixed-precision training often keeps a higher-precision master copy of the weights for the optimiser step; this update is typically performed in FP32, and using higher precision there typically has little measurable performance impact in practice~\citep{micikevicius2018mixed,mellempudi2019mixed}.

Second-order optimisers and L-BFGS: Quasi-Newton optimisers are commonly used in PINN training because PINN objectives are often highly ill-conditioned, making purely first-order optimisation difficult~\citep{URBAN2025113656}. L-BFGS is a limited-memory quasi-Newton optimiser~\citep{Nocedal1980UpdatingQM,liuLimitedMemoryBFGS1989}, which approximates the inverse Hessian metric based on parameter and loss update information. It is preferred over full BFGS/Newton-like dense methods because it is limited-memory and designed for large-scale optimisation.

Optimisation difficulty and conditioning in PINNs: The training difficulty of PINNs is closely tied to the optimisation landscape induced by the PDE residual and constraint terms. Early studies identified characteristic failure modes in which PINNs fit the training objective poorly on nontrivial PDEs, for instance, equations with sharp gradients or localised solution transitions~\citep{krishnapriyan2021characterizing,GIE2026116918}. Moreover, the differential-operator-based regularisation used in PINNs can substantially worsen conditioning~\citep{krishnapriyan2021characterizing}. More recent work has analysed these optimisation difficulties through curvature-related properties of the loss landscape, such as the Hessian spectrum and condition number~\citep{CAO2025113494,rathore2024challenges}. These observations motivate the use of curvature-aware signals for adapting numerical precision during PINN training.

\section{Methods}
This section outlines the proposed methodology and evaluation framework. We first introduce PINNs for forward problems, then present the proposed curvature-aware precision controller for adaptive numerical precision. Finally, we describe the network architectures and evaluation criteria used to assess accuracy and computational efficiency.

\begin{figure}
    \centering
    \includegraphics[width=0.99\linewidth]{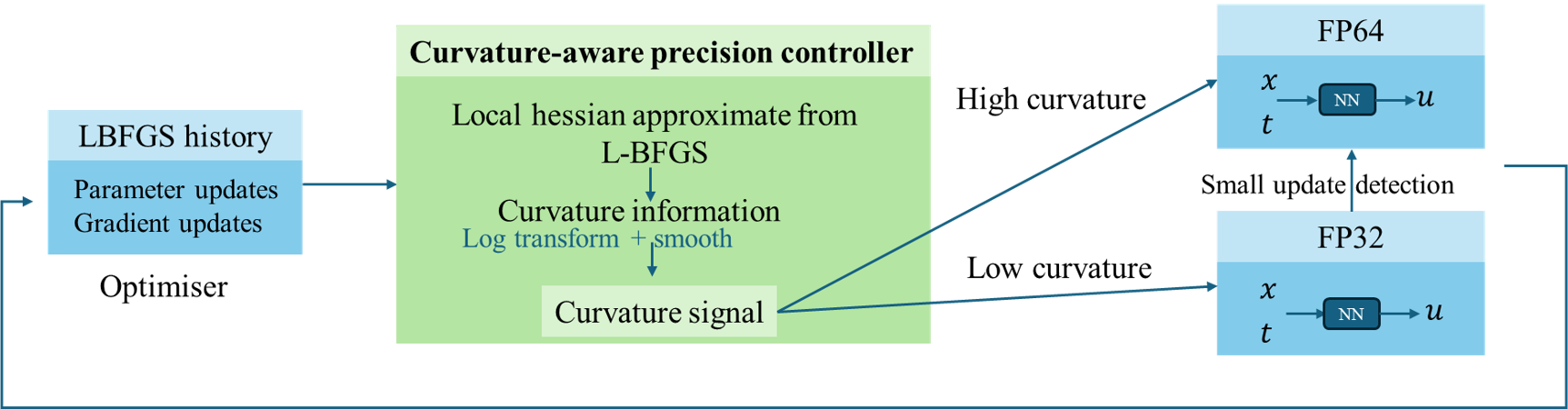}
    \caption{Curvature-aware adaptive precision for PINN. The proposed controller uses parameter and gradient update information to calculate a curvature signal for precision adaptation. The high curvature phase trains in double precision, and the low curvature phase trains in single precision.}
    \label{fig:mainfigure}
\end{figure}

\subsection{PINN for forward problems}\label{sec:PINN}
The PINN structure used in this work is considered as a neural network-based solver for forward differential-equation problems~\citep{RAISSI2019686}. Let \(x \in \Omega \subset \mathbb{R}^D\) denote the spatial variable, where \(\Omega\) is the spatial domain and \(D\) is the spatial dimension, and let \(t \in [0,T]\) denote the time variable over the time interval with final time \(T\). The governing equation is written in residual form as
\begin{equation}\label{eq:pinn_loss_residual_format}
    \mathcal{N}[u] = 0, \qquad (x,t) \in \Omega \times [0,T],
\end{equation}
where \(u := u(x,t)\) is the unknown solution field and \(\mathcal{N}\) denotes the differential operator, which is assumed to be known for the forward problem in this study.

In the PINN structure, the neural-network approximation to \(u(x,t)\) is denoted by \(u_\theta(x,t)\), where \(\theta\) denotes the trainable network parameters. The model is trained by minimising the total loss
\begin{align} \label{eq:pinn_loss}
    \mathcal{L} = \lambda_\mathrm{res} \mathcal{L}_\mathrm{res} + \lambda_B \mathcal{L}_B + \lambda_I \mathcal{L}_I,
\end{align}
where \(\mathcal{L}_\mathrm{res}\), \(\mathcal{L}_B\), and \(\mathcal{L}_I\) denote the residual, boundary-condition, and initial-condition losses, respectively, and \(\lambda_\mathrm{res}\), \(\lambda_B\), and \(\lambda_I\) are their corresponding weights.

The residual loss is defined as
\begin{align}\label{eq:pinn_res_loss}
    \mathcal{L}_{\mathrm{res}} = \frac{1}{N_{\mathrm{res}}} \sum_{i=1}^{N_{\mathrm{res}}} \left|\mathcal{N}[u_\theta(x_i^{\mathrm{res}},t_i^{\mathrm{res}})]\right|^2,
\end{align}
where \(\{(x_i^{\mathrm{res}},t_i^{\mathrm{res}})\}_{i=1}^{N_{\mathrm{res}}}\) are collocation points sampled from the space-time domain. The boundary and initial-condition losses are problem-dependent and are specified later for each benchmark equation.

We use the L-BFGS optimiser to train the PINN models. L-BFGS is a quasi-Newton method that approximates curvature information without explicitly forming or storing the full Hessian matrix. Instead, it constructs an implicit inverse-Hessian approximation from stored parameter-update and gradient-update pairs. Let
\begin{equation}
\begin{aligned}\label{eq:lbfgs_s_y}
    s_k &= \theta_{k+1} - \theta_k,\\
    y_k &= \nabla_\theta \mathcal{L}(\theta_{k+1}) - \nabla_\theta \mathcal{L}(\theta_k),
\end{aligned}
\end{equation}
denote the parameter differences (\(s_k\))and gradient differences (\(y_k\)) at iteration \(k\), where \(\theta_k\) and \(\nabla_\theta \mathcal{L}(\theta_k)\) are the network parameters and the loss gradient with respect to \(\theta\) at iteration \(k\). These stored curvature pairs\((s_k,y_k)\)  are then used to compute a quasi-Newton search direction for updating the neural-network parameters during training.

\subsection{Proposed Curvature-aware precision controller}\label{sec:curvature_aware}

To distinguish between well-conditioned and ill-conditioned phases of training, we use a loss curvature proxy derived from the history stored by the L-BFGS optimiser~\citep{Nocedal1980UpdatingQM,liuLimitedMemoryBFGS1989}. The curvature pairs maintained by L-BFGS to construct its quasi-Newton update can therefore be reused for curvature signals with negligible additional overhead.

In L-BFGS, the scalar \(s_k^\top y_k\) contains curvature information and is used to maintain the positive-definiteness of the quasi-Newton update. From the secant relation \(B_{k+1}s_k = y_k\), where \(B_{k+1}\) denotes the quasi-Newton approximation of the Hessian, we define
\begin{align}\label{eq:kappa_k}
    \kappa_k = \frac{s_k^\top B_{k+1} s_k}{s_k^\top s_k} = \frac{s_k^\top y_k}{s_k^\top s_k}.
\end{align}

The term \(\kappa_k\) can be interpreted as the Rayleigh quotient~\citep{doi:10.1137/1.9781611977165.ch5} of the local Hessian approximation along direction \(s_k\), representing directional curvature. This quantity reflects curvature along recent parameter update directions and can be computed without explicitly forming \(B_{k+1}\).

For a symmetric positive-definite Hessian, the condition number is equivalent to the ratio between the maximum and minimum Rayleigh quotients over all nonzero directions~\citep{Boyd_Vandenberghe_2004}. Following this form, we define a conditioning proxy using the valid curvature pairs stored in the current L-BFGS memory
\begin{align}\label{eq:consition_number_proxy}
    \hat{\kappa}_{\mathrm{cond}} =
    \frac{\max_{(s_k,y_k) \in \mathcal{H}_j} \kappa_k}
    {\min_{(s_k,y_k) \in \mathcal{H}_j} \kappa_k},
\end{align}
where \(\mathcal{H}_j\) denotes the set of valid stored L-BFGS curvature pairs available at L-BFGS optimisation step \(j\). We treat a curvature pair as valid when it satisfies the positive-curvature condition \(s_k^\top y_k > 0\), the denominator \(s_k^\top s_k\) is numerically nonzero, and the resulting curvature value is finite. The resulting quantity measures the spread of secant-based directional curvature estimates over the recent L-BFGS update directions, and serves as a lightweight proxy for local conditioning.

To stabilise this local condition signal, a log transform and exponential moving average smoothing is applied~\citep{stevensLogarithmicTransformation1946,264b1310-db27-36ce-b68f-3e439b03c2b5}
\begin{equation}
\begin{aligned}\label{eq:smooth_curvature}
    z_j &= \log_{10}(\hat{\kappa}_{\mathrm{cond}} + \varepsilon),\\
    \tilde{z}_j &= \alpha \tilde{z}_{j-1} + (1-\alpha) z_j,
\end{aligned}
\end{equation}
where \(\varepsilon=10^{-12}\) is a small constant introduced for numerical stability, and we chose \(\alpha = 0.9\) so that the smoothed proxy emphasises persistent trends in conditioning while reducing sensitivity to short-term fluctuations caused by fluctuations in curvature estimates~\cite{doi:10.1137/20M1373190}. Furthermore a conditioning proxy slope is defined as
\begin{align}\label{eq:condition_slope}
    \Delta \tilde{z}_j = \tilde{z}_j - \tilde{z}_{j-1}.
\end{align}

Our proposed precision controller uses both \(\tilde{z}_j\) and \(\Delta \tilde{z}_j\) to decide when to switch between FP32 and FP64 (Figure~\ref{fig:mainfigure}). FP32 is used only when \(\tilde{z}_j\) is below a prescribed threshold \(\tau_z\) and \(|\Delta \tilde{z}_j|\) remains small. This indicates a stable and well-conditioned phase of training. The controller switches back to FP64 when \(\tilde{z}_j\) becomes large, \(\Delta \tilde{z}_j\) increases rapidly, or relative parameter updates in FP32 become extremely small. To prevent rapid oscillation between precision states, switching decisions are evaluated every 10 L-BFGS training steps.

\begin{algorithm}[!htbp]
\caption{Curvature-aware dynamic precision controller}
\label{alg:curvature_controller}
\begin{algorithmic}[1]
\Require L-BFGS optimiser, initial precision FP64, curvature threshold \(\tau_z\), maximum step count \(J_{\max}\)
\For{L-BFGS optimisation step \(j=1,\ldots,J_{\max}\)}
    \State Perform one L-BFGS update using the current precision
    \If{early-stopping criterion is satisfied}
        \State \textbf{break}
    \EndIf
    \If{\(j \bmod 10 = 0\)}
        \State Extract valid L-BFGS curvature pairs, save in \(\mathcal{H}_j\)
        \If{\(\mathcal{H}_j\) contains at least two valid curvature pairs}
            \State Compute \(\hat{\kappa}_{\mathrm{cond}}\), \(z_j\), \(\tilde{z}_j\), and \(\Delta \tilde{z}_j\)
            \If{current precision is FP64, \(\tilde{z}_j<\tau_z\), and \(|\Delta \tilde{z}_j|<0.02\)}
                \State Switch to FP32
            \ElsIf{current precision is FP32 and \((\tilde{z}_j>\tau_z\) or \(\Delta \tilde{z}_j>0.03\) or FP32 tiny updates are detected)}
                \State Switch to FP64
            \EndIf
        \EndIf
    \EndIf
\EndFor
\end{algorithmic}

\vspace{0.4em}
\begin{minipage}{\linewidth}
\footnotesize
\textit{Note.} In practice, \(J_{\max}\) is set to a sufficiently large value, and training terminates through early stopping. The early-stopping criterion is satisfied when the total training loss \(\mathcal{L}\) does not improve by more than \(10^{-7}\) for 50 consecutive training steps after 100 warm-up steps. FP32 tiny updates are detected when the maximum relative parameter update is below \(10^{-8}\).
\end{minipage}

\end{algorithm}

Prior work has shown that training difficulty in PINNs is strongly affected by the differential operator appearing in the residual term and by the resulting loss-landscape conditioning, while optimisation behaviour also depends on the choice of training algorithm~\citep{rathore2024challenges,krishnapriyan2021characterizing}. More broadly, the PINN literature reports equation-dependent training pathologies, including imbalance between residual and boundary terms and difficulties specific to time-dependent problems, and commonly evaluates methods under problem-dependent numerical setups~\citep{pmlr-v202-muller23b}. In this study, we use the same controller across PDEs but treat the threshold \(\tau_z\) applied to the smoothed curvature proxy \(\tilde{z}_j\) as a problem-dependent hyperparameter and allow it to vary by equation. Moreover, we include an ablation study to examine the sensitivity of performance to different values of \(\tau_z\). The curvature-aware controller is summarised in Algorithm~\ref{alg:curvature_controller}.

\subsection{Network architectures and evaluation}\label{sec:Network_architectures_and_evaluation}
The proposed approach is evaluated on four representative neural-network architectures: a standard MLP-based PINN baseline~\citep{RAISSI2019686} and three recently proposed advancements, namely PINNMamba~\citep{xu2025sub}, PINNsFormer~\citep{zhao2023pinnsformer}, and KAN~\citep{liukan}. The MLP serves as the conventional PINN architecture, while PINNMamba and PINNsFormer are designed to improve the propagation of temporal or initial-condition information in PINNs, which is a known difficulty in time-dependent PDE benchmarks. KAN is included as an alternative neural network structure to MLP-based PINN, as recent works suggests it improved accuracy for several forward and inverse PDE problems~\citep{wang2025kolmogorov}.

All models are trained on a single NVIDIA A100 GPU. To evaluate the performance of different approaches, we report the relative root mean squared error (rRMSE), the relative mean absolute error (rMAE), and the total training wall-clock time.
PINN predictions \(u_\theta\) are evaluated on a fixed benchmark grid with \(N\) evaluation points. For each benchmark equation, the collocation points used for training are sampled from the same prescribed grid, so all precision strategies are compared under identical spatial-temporal sampling. We define
\begin{align}\label{eq:rmse_rmae}
    \mathrm{rRMSE} &= \sqrt{\frac{\sum_{i=1}^N \left(u(x_i,t_i) - u_\theta(x_i,t_i)\right)^2}{\sum_{i=1}^N u(x_i,t_i)^2}},\\
    \mathrm{rMAE} &= \frac{\sum_{i=1}^N \left|u(x_i,t_i) - u_\theta(x_i,t_i)\right|}{\sum_{i=1}^N \left|u(x_i,t_i)\right|}.
\end{align}

The same stopping criterion is applied to all PINN models. After an initial 100-step warm-up, training is terminated when the total training loss \(\mathcal{L}\) does not improve by more than \(10^{-7}\) for more than 50 consecutive outer L-BFGS optimiser steps.

\section{Experiments and results}\label{sec:result_discussion}
This section presents the experimental evaluation of the proposed curvature-aware dynamic precision strategy. We first describe the experimental setup, then compare its performance against fixed-precision baselines. We further examine the effects of network architecture, the generalisation of dynamic precision, and the relationship between curvature signals and training dynamics. Finally, we demonstrate the method on an irradiance-driven plant-growth ODE and discuss its limitations and future directions.

\subsection{Setup}
The proposed approach is evaluated on four PINN failure-mode benchmark equations and one ordinary differential equation (ODE) example that represents a real-life application. The overall experiments are summarised in Table~\ref{tab:experimental_suite_dynamic_precision}.

PINN failure-mode benchmark equations: The four one-dimensional benchmark equations are the \textit{Convection} equation, \textit{Reaction} equation, \textit{Wave} equation, and \textit{Allen--Cahn} equation. These equations are commonly used for analysing the failure modes of PINNs~\citep{krishnapriyan2021characterizing,xu2025fp64}, and are defined as follows.

The 1D \textit{Convection} equation used in this study describes the transport of a scalar quantity at a constant speed of 50, representing a challenging high-speed transport benchmark in which PINNs can suffer from propagation failures associated with highly imbalanced PDE residual fields~\citep{daw2023mitigating}. The problem is defined on the periodic spatial domain \(x \in [0,2\pi]\) and the time interval \(t \in [0,1]\). The initial condition is \(u(x,0)=\sin(x)\), and periodic boundary conditions are imposed at the two spatial boundaries as defined
\begin{equation}
\label{eq:convection}
\begin{aligned}
&\dfrac{\partial u}{\partial t} + 50 \dfrac{\partial u}{\partial x} = 0, \\
&u(x,0) = \sin(x), \\
&u(0,t) = u(2\pi,t).
\end{aligned}
\end{equation}

\begin{table}[t]
\centering
\caption{Overview of the performed experiments.}
\label{tab:experimental_suite_dynamic_precision}
\small
\begin{tabularx}{\linewidth}{@{}
>{\centering\arraybackslash}p{0.24\linewidth}
>{\centering\arraybackslash}p{0.26\linewidth}
>{\centering\arraybackslash}X
>{\centering\arraybackslash}p{0.18\linewidth}
@{}}
\toprule
\textbf{Experiment} & \textbf{Equations} & \textbf{Network architecture} & \textbf{Results} \\
\midrule

Dynamic approach and fixed-precision comparison &
\textit{Convection}, \textit{Reaction}, \textit{Wave}, \textit{Allen--Cahn} &
Vanilla PINN &
Table~\ref{tab:main_comparison} \\

Threshold ablation &
\textit{Convection}, \textit{Reaction}, \textit{Wave}, \textit{Allen--Cahn} &
Vanilla PINN &
Table~\ref{tab:different_threshold_ablation} \\

\midrule

Architecture comparison &
\textit{Convection}, \textit{Reaction}, \textit{Wave}, \textit{Allen--Cahn} &
Vanilla PINN, PINNsFormer, PINNMamba, KAN &
Table~\ref{tab:different_model} \\

Wider shallower MLP &
\textit{Convection}, \textit{Reaction}, \textit{Wave}, \textit{Allen--Cahn} &
Three-layer vanilla PINN &
Table~\ref{tab:three_layer_mlp} \\

Dynamic-approach generalisation &
PINNsFormer: \textit{Convection}, \textit{Reaction}, \textit{Wave}, Allen--Cahn; PINNMamba: \textit{Wave}, \textit{Convection} &
PINNsFormer, PINNMamba &
Table~\ref{tab:curvature_aware_PINNsFormer}; Table~\ref{tab:wave_PINNmamba} \\

\midrule

Application example &
Irradiance-driven plant-growth ordinary differential equation &
Vanilla PINN &
Table~\ref{tab:irradiance_ode} \\

\bottomrule
\end{tabularx}
\end{table}

The \textit{Reaction} equation describes local nonlinear growth through a logistic-type reaction term. It is included as a PINN failure-mode benchmark because previous work has shown that reaction-dominated problems with larger reaction coefficients can be difficult for vanilla PINNs to optimise. The spatial dependence is introduced through the initial condition, while the solution evolves over the same periodic spatial domain \(x \in [0,2\pi]\) and time interval \(t \in [0,1]\). The initial condition is a Gaussian profile centred at \(x=\pi\), and periodic boundary conditions are imposed
\begin{equation}
\label{eq:reaction}
\begin{aligned}
&\dfrac{\partial u}{\partial t} - 5 u(1-u) = 0, \\
&u(x,0) = \exp\!\Big(-\dfrac{(x-\pi)^2}{2(\pi/4)^2}\Big), \\
&u(0,t) = u(2\pi,t).
\end{aligned}
\end{equation}


The \textit{Wave} equation is a second-order hyperbolic PDE that describes wave propagation. In this benchmark, the solution is defined on the spatial domain \(x \in [0,1]\) and the time interval \(t \in [0,1]\). The initial-condition loss contains both a displacement term and a velocity term, while the boundary-condition loss enforces homogeneous Dirichlet conditions at \(x=0\) and \(x=1\)
\begin{equation}
\label{eq:wave}
\begin{aligned}
&\dfrac{\partial^2 u}{\partial t^2} - 4\dfrac{\partial^2 u}{\partial x^2} = 0, \\
&u(x,0) = \sin(\pi x) + \tfrac{1}{2}\sin(3\pi x), \\
&\dfrac{\partial u}{\partial t}(x,0) = 0, \\
&u(0,t) = u(1,t) = 0.
\end{aligned}
\end{equation}

The \textit{Allen--Cahn} equation is a nonlinear equation. In this benchmark, the small diffusion coefficient and nonlinear cubic term make the equation challenging for PINN training. The problem is defined on the spatial domain \(x \in [-1,1]\) and the time interval \(t \in [0,1]\). The boundary-condition loss includes both value matching and derivative matching at the two boundaries

\begin{equation}
\label{eq:allencahn}
\begin{aligned}
&\dfrac{\partial u}{\partial t} - 0.0001\,\dfrac{\partial^2 u}{\partial x^2} + 5u^3 - 5u = 0, \\
&u(x,0) = x^2\cos(\pi x), \\
&u(-1,t) = u(1,t), \\
&\dfrac{\partial u}{\partial x}(-1,t) = \dfrac{\partial u}{\partial x}(1,t).
\end{aligned}
\end{equation}

To ensure comparability on the four PINN failure mode benchmark equations, we follow the model configurations, loss definitions, collocation-point settings, and evaluation approach of~\citet{xu2025fp64}, using a \(101 \times 101\) collocation grid for each PDE experiment. The PINN predictions are evaluated against analytical solutions for the \textit{convection}, \textit{reaction}, and \textit{wave} equations, and against the high-resolution spectral approximation~\cite{kopriva2009implementing,RAISSI2019686,xu2025fp64} for the \textit{Allen--Cahn} equation, where no closed-form analytical solution is available.

\begin{figure}
    \centering
    \includegraphics[width=0.7\linewidth]{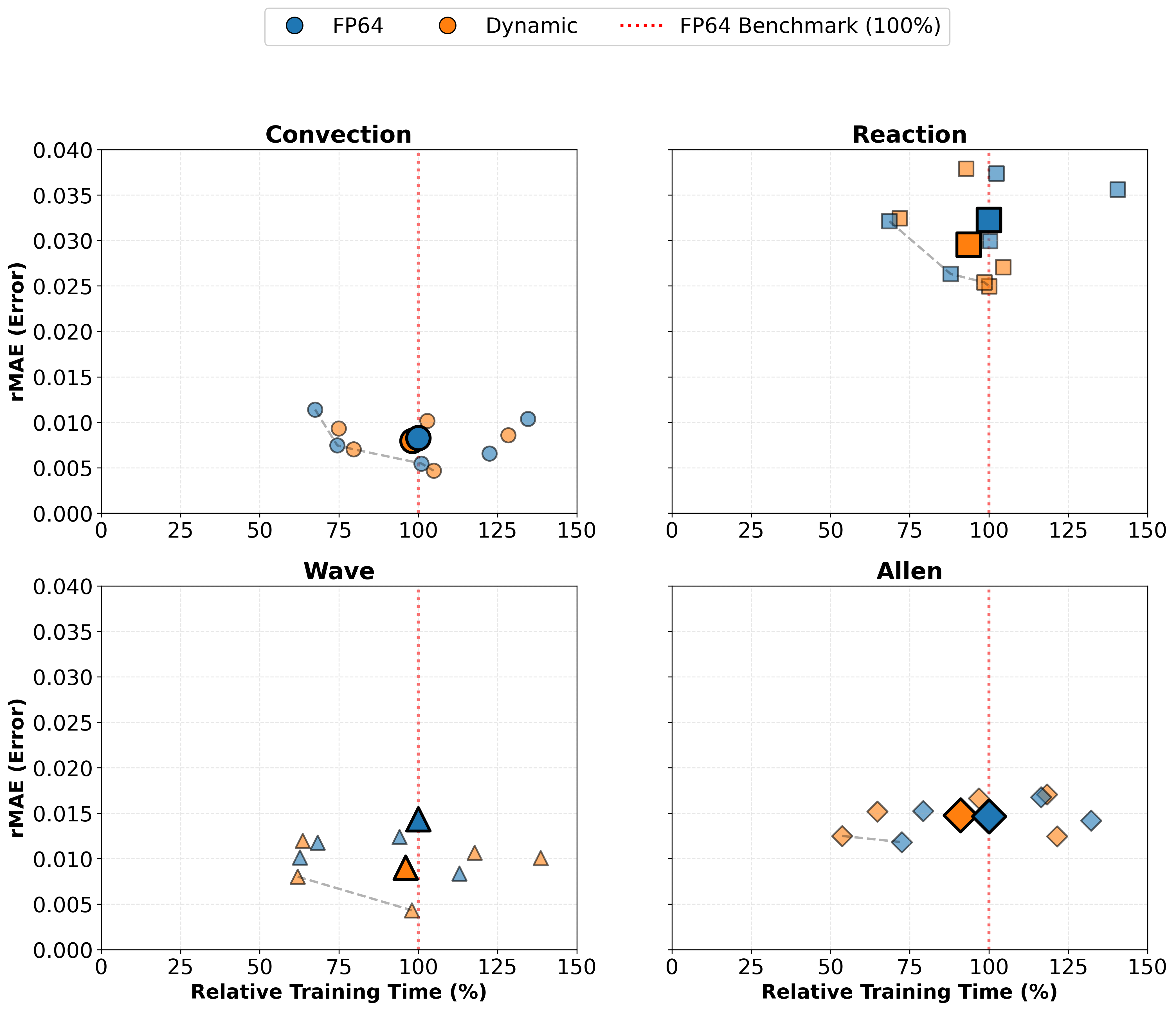}
    \caption{Accuracy and speed improvement for four PINN failure mode benchmark equations. The plot compares the proposed curvature-aware dynamic-precision method with the FP64 baseline across random seeds. Larger markers denote the average over all seeds. The red vertical line marks the FP64 training-time baseline, and the grey dashed line indicates the Pareto frontier formed by the non-dominated runs in each panel. Points closer to the bottom-left corner indicate better speed--accuracy trade-offs.}
    \label{fig:Speed_accuracy}
\end{figure}

Furthermore, we consider an ODE system motivated by plant growth. Following the irradiance model structure of~\citet{10.3389/fpls.2023.1172359}, we assume that the biomass dynamics are governed by a logistic-type ODE affected by irradiance. We use a modified form of the equation, defined as
\begin{equation}
\label{eq:irradiance}
\begin{split}
\frac{du(t;A)}{dt} &= \left(r + S(t;A)\right)u(t;A)
\left(1-\frac{u(t;A)}{K}\right),\\
S(t;A) &= A\sin\!\left(4\pi(t+0.75)\right),
\end{split}
\end{equation}
where \(A\) controls the amplitude of the seasonal driver, and \(S(t;A)\) represents a simulated irradiance effect on the biomass growth rate. The parametrisation is modified relative to the original paper because time is normalised to \(t\in[0,1]\). The ground-truth solution is obtained numerically using the RK45 solver~\citep{dormandFamilyEmbeddedRungeKutta1980}. In our PINN implementation, the network takes \(t\) and \(S(t;A)\) as inputs and learns the corresponding solution family \(u_\theta(t,A)\) for \(t\in[0,1]\) and \(A\in[-1,1]\), with initial condition \(u_\theta(0,A)=0.05\). For the irradiance ODE, we sample \(10{,}100\) collocation points over the normalised time interval and evaluate the solution family using five values of \(A\) to represent different irradiance amplitudes.

\subsection{Performance comparison against fixed precision}
The proposed method is compared with fixed precision cases for the same problem under same conditions. In particular, for fixed precision, FP32 and FP64 training are carried out for the full problem. As shown in Table~\ref{tab:main_comparison} and Figure~\ref{fig:Speed_accuracy}, the proposed dynamic-precision method improves the average speed--accuracy trade-off relative to FP64 across the tested equations. The dynamic-precision results in the main table use the selected threshold setting for each equation, while the results for the other tested \(\tilde{z}_j\) threshold values are reported in Table~\ref{tab:different_threshold_ablation}. The proposed curvature-aware dynamic approach shows greater improvement for the \textit{Reaction} and \textit{Wave} equations with vanilla PINN, in both speed and accuracy, while \textit{Allen--Cahn} shows near-equal accuracy with modest training speed improvement. Pure FP32 is substantially faster but often fails to converge to an acceptable solution, producing errors that are one to two orders of magnitude larger. 
\begin{center}
\captionsetup{hypcap=false}
\captionof{table}{Average results over 5 random seeds for the vanilla PINN. `Speedup vs FP64' is computed as the ratio between the average FP64 training time and the average training time of the corresponding method for the same benchmark. Dynamic-precision results are highlighted in \textcolor{blue}{blue}, and the lowest error is \textbf{bold}.}
\label{tab:main_comparison}
\begin{adjustbox}{max width=0.99\textwidth}
\begin{tabular}{cccccc}
\toprule
Equation & Approach & Avg Time (s) & rRMSE & rMAE & Speedup vs FP64 \\
\midrule
\multirow{3}{*}{\textit{Convection} (Eq.~\ref{eq:convection})}
& \textcolor{blue}{dynamic} 
& \textcolor{blue}{\(\mathbf{1,071.94} \pm \mathbf{235.94}\)} 
& \textcolor{blue}{\(\mathbf{0.0091} \pm \mathbf{0.0024}\)} 
& \textcolor{blue}{\(\mathbf{0.0079} \pm \mathbf{0.0022}\)} 
& \textcolor{blue}{\(\mathbf{1.02}\times\)} \\
& FP32 & \(271.97 \pm 165.15\) & \(0.8776 \pm 0.2887\) & \(0.7906 \pm 0.2662\) & \(4.02\times\) \\
& FP64 & \(1,092.68 \pm 319.15\) & \(0.0108 \pm 0.0051\) & \(0.0083 \pm 0.0025\) & \(1.00\times\) \\

\midrule
\multirow{3}{*}{\textit{Reaction} (Eq.~\ref{eq:reaction})}
& \textcolor{blue}{dynamic} 
& \textcolor{blue}{\(662.94 \pm 91.11\)} 
& \textcolor{blue}{\(\mathbf{0.0531} \pm 0.0101\)} 
& \textcolor{blue}{\(\mathbf{0.0295} \pm 0.0055\)} 
& \textcolor{blue}{\(\mathbf{1.07}\times\)} \\
& FP32 & \(65.95 \pm 16.39\) & \(0.9779 \pm 0.0015\) & \(0.9770 \pm 0.0010\) & \(10.74\times\) \\
& FP64 & \(708.51 \pm 187.09\) & \(0.0578 \pm 0.0075\) & \(0.0323 \pm 0.0044\) & \(1.00\times\) \\

\midrule
\multirow{3}{*}{\textit{Wave} (Eq.~\ref{eq:wave})}
& \textcolor{blue}{dynamic} 
& \textcolor{blue}{\(1,283.06 \pm 448.48\)} 
& \textcolor{blue}{\(\mathbf{0.0094} \pm 0.0033\)} 
& \textcolor{blue}{\(\mathbf{0.0090} \pm 0.0030\)} 
& \textcolor{blue}{\(\mathbf{1.04}\times\)} \\
& FP32 & \(483.85 \pm 118.35\) & \(0.0407 \pm 0.0421\) & \(0.0374 \pm 0.0360\) & \(2.76\times\) \\
& FP64 & \(1,336.86 \pm 537.72\) & \(0.0147 \pm 0.0087\) & \(0.0143 \pm 0.0084\) & \(1.00\times\) \\

\midrule
\multirow{3}{*}{\textit{Allen--Cahn} (Eq.~\ref{eq:allencahn})}
& \textcolor{blue}{dynamic} 
& \textcolor{blue}{\(3,375.30 \pm 1,140.70\)} 
& \textcolor{blue}{\(\mathbf{0.0495} \pm 0.0097\)} 
& \textcolor{blue}{\(0.0148 \pm 0.0022\)} 
& \textcolor{blue}{\(\mathbf{1.10}\times\)} \\
& FP32 & \(53.08 \pm 37.24\) & \(0.9239 \pm 0.0629\) & \(0.8588 \pm 0.1960\) & \(69.88\times\) \\
& FP64 & \(3,709.20 \pm 926.90\) & \(0.0503 \pm 0.0067\) & \(\mathbf{0.0147} \pm 0.0018\) & \(1.00\times\) \\
\bottomrule
\end{tabular}
\end{adjustbox}
\end{center}

The curvature information also varies across different PDEs and is related to their optimisation behaviour (Figure~\ref{fig:log_proxy}). For the \textit{Reaction} and \textit{Wave} equations, we observe relatively small and smooth curvature during training, while the \textit{Convection} and \textit{Allen--Cahn} equations show higher curvature and more fluctuation. This is also linked to the improvement from dynamic precision, where the \textit{Wave} and \textit{Reaction} equations show larger improvements in speed--accuracy trade-off (Figure~\ref{fig:Speed_accuracy}), while more frequent switching is noticed in the \textit{Convection} and \textit{Allen--Cahn} cases, as shown in Figure~\ref{fig:precision_switching_example}.

In principle, dynamic precision uses the same model class and loss function as pure FP64, which should result in the same solution space. However, we notice that our approach can sometimes reach a lower prediction error in practice. A likely reason is that switching precision acts as a numerical perturbation that changes the optimisation path (Figure~\ref{fig:log_proxy}). Different training strategies produce different curvature histories, and the dynamic-precision trajectory does not simply replicate the FP64 trajectory, especially near the end of training. The L-BFGS optimiser uses curvature information estimated from recent parameter and gradient differences to control the optimisation path, and switching precision may slightly change the loss-update history near convergence, allowing the optimiser to terminate at a different point with a lower final error.
\begin{center}
\begin{minipage}{\linewidth}
\centering
\newcommand{\panelheight}{0.18\textheight}

\begin{minipage}{0.50\linewidth}
    \centering
    \includegraphics[height=\panelheight]{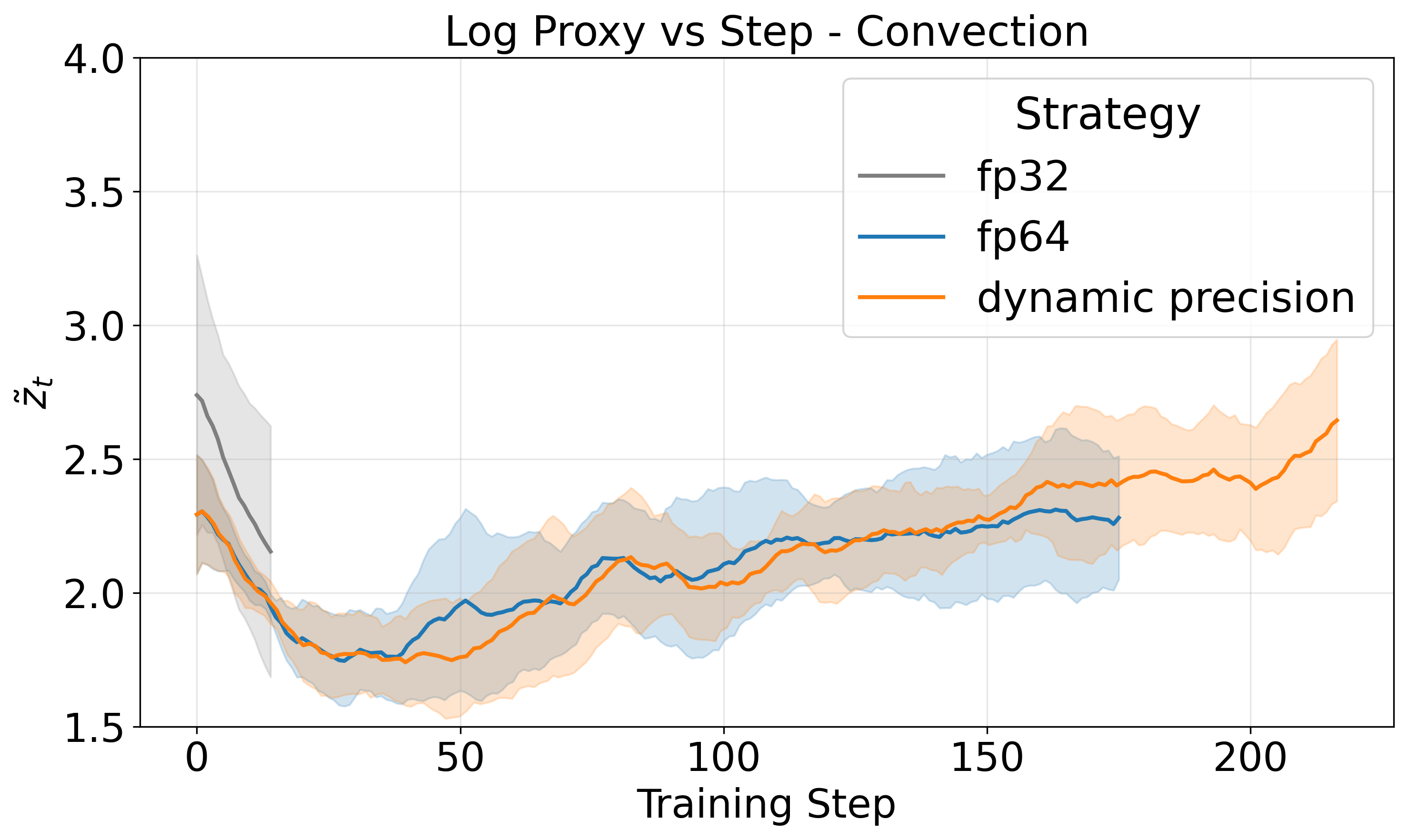}\\[-0.35em]
    {\small (a) \textit{Convection}}
\end{minipage}%
\begin{minipage}{0.50\linewidth}
    \centering
    \includegraphics[height=\panelheight]{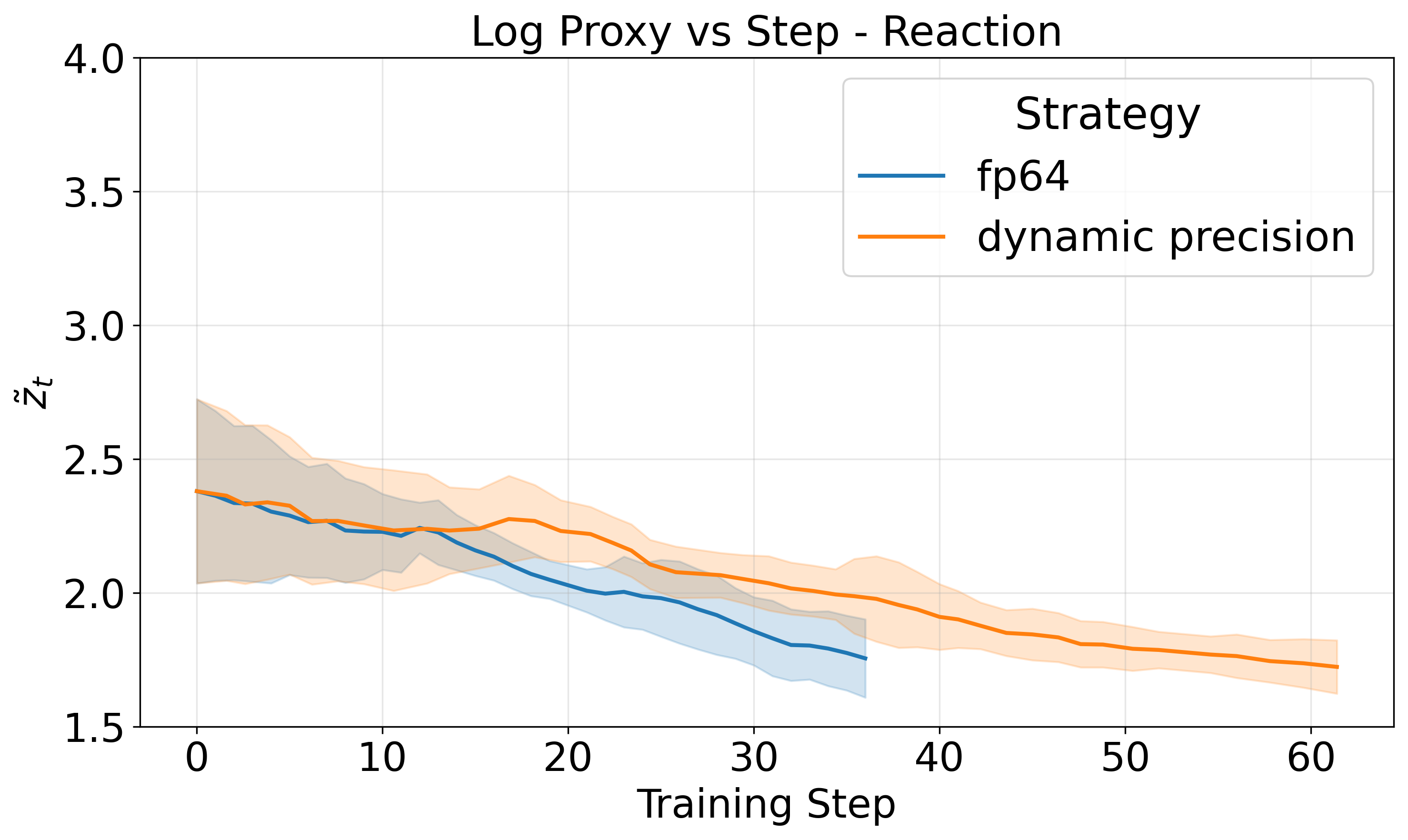}\\[-0.35em]
    {\small (b) \textit{Reaction}}
\end{minipage}\\[-0.2em]
\begin{minipage}{0.50\linewidth}
    \centering
    \includegraphics[height=\panelheight]{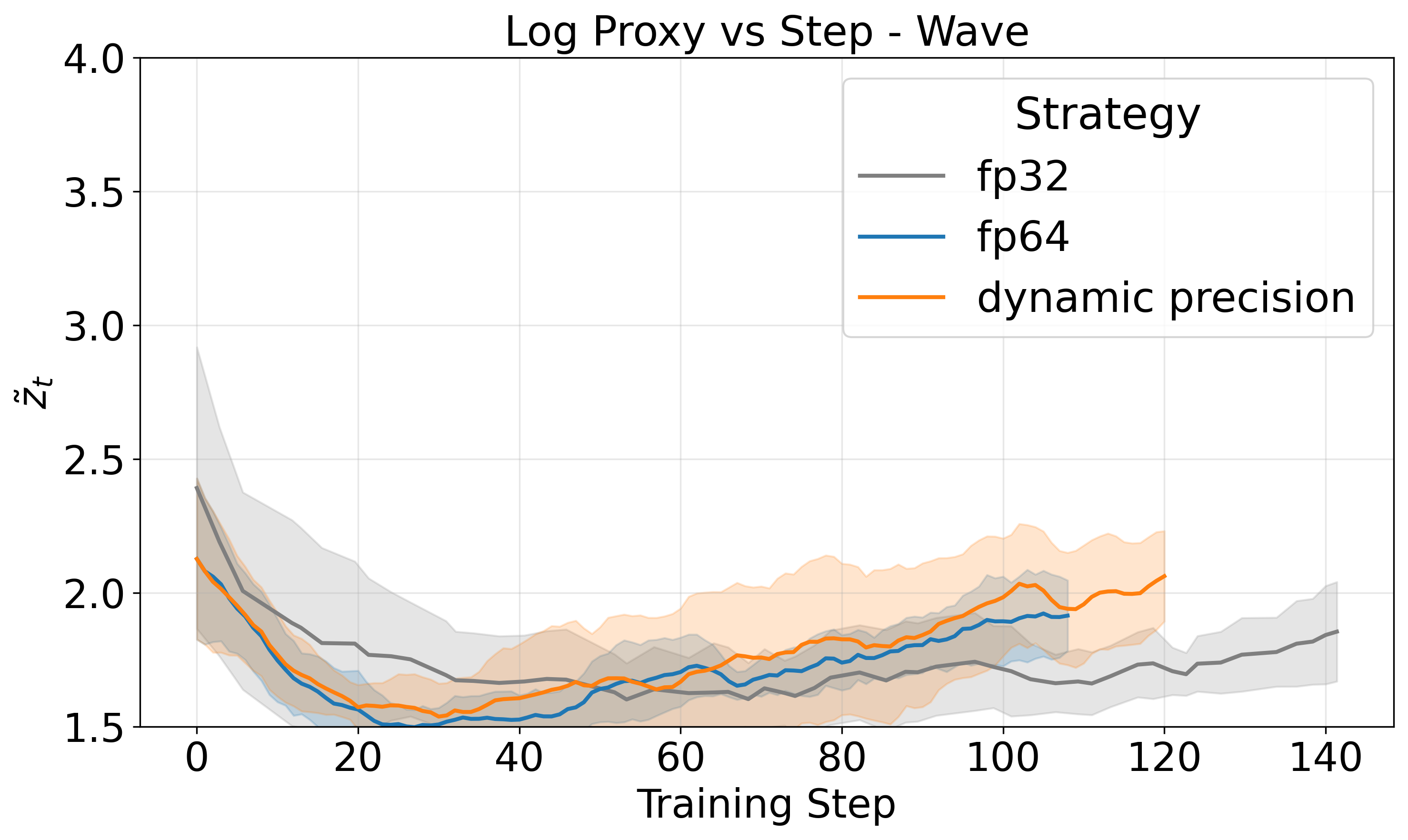}\\[-0.35em]
    {\small (c) \textit{Wave}}
\end{minipage}%
\begin{minipage}{0.50\linewidth}
    \centering
    \includegraphics[height=\panelheight]{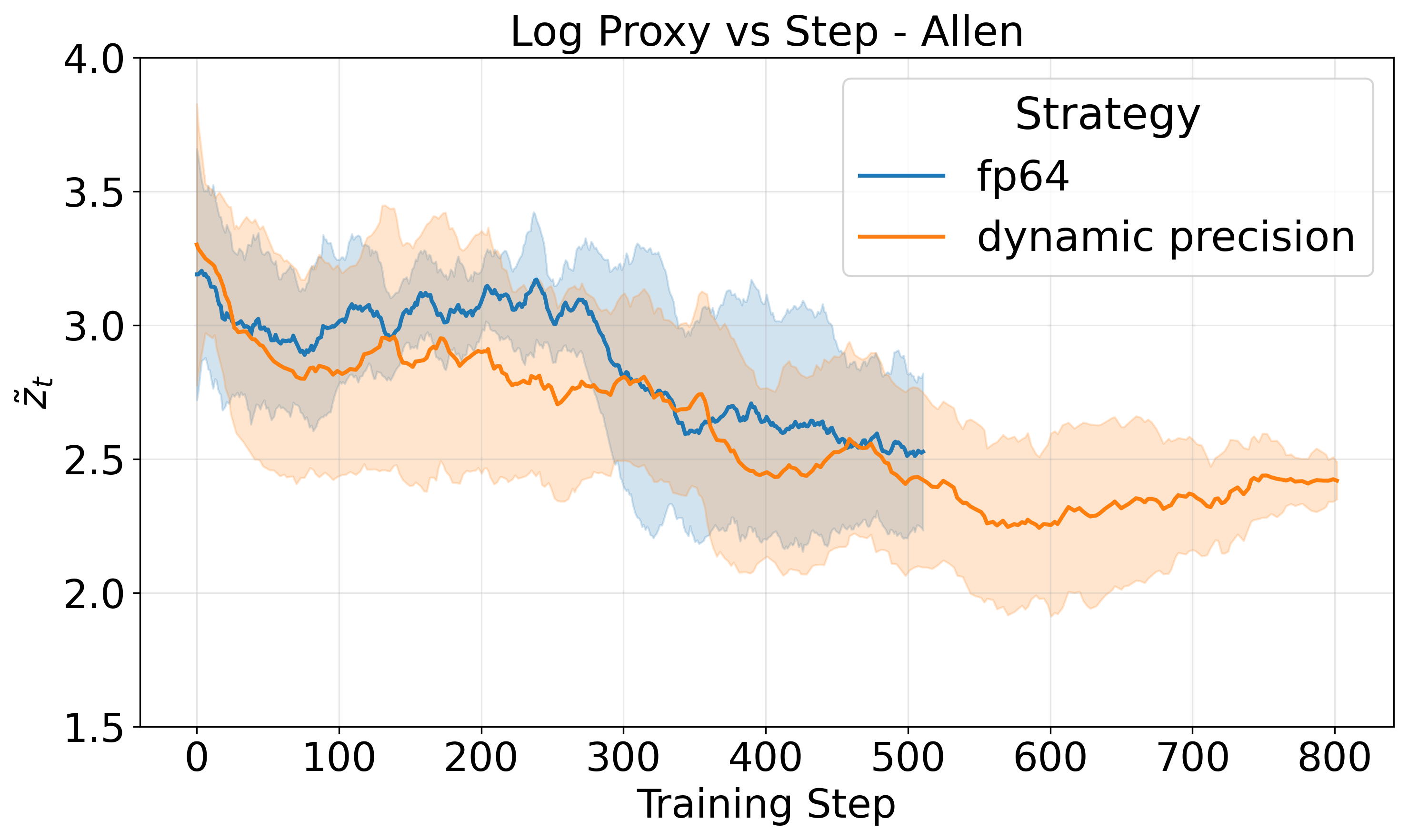}\\[-0.35em]
    {\small (d) Allen--Cahn}
\end{minipage}
\captionsetup{hypcap=false}
\captionof{figure}{Curvature-aware signal \(\tilde{z}_j\) during training for different precision strategies and benchmark equations. Solid lines show the mean over random seeds, and shaded regions indicate the standard deviation where at least three runs are available. The trajectory length may differ across strategies because runs terminate at different training steps.}
\label{fig:log_proxy}
\end{minipage}
\end{center}

\subsection{Architecture effects and generalisation of dynamic precision}\label{sec:different_structure}

\begin{center}
    \centering
    \captionsetup{hypcap=false}
    \captionof{table}{PINNformer, PINNmamba and KAN baseline result compared with dynamic precision with MLP. The result with FP32 is not always converged, so we only calculated the result with the seeds that converged (rRMSE<0.3) unless all five seeds have high errors (Convergence rate = 0\% noted with *). }
    \label{tab:different_model}
    \begin{adjustbox}{max width=0.99\textwidth}
    \begin{tabular}{cccccccc}
\toprule
Equation & approach & Avg Time (s) & rRMSE & rMAE &Convergence rate \\
\midrule
\multirow{4}{*}{\textit{Convection} (Eq.~\ref{eq:convection})} 
& PINNsFormer & 1136.4 $\pm$ 109.0 & 0.0209 $\pm$ 0.0079 & 0.0192 $\pm$ 0.0072 & $60\%$ \\
& PINNmamba & \(168.19 \pm 111.32\) & \(1.3330 \pm 0.3474\) & \(1.2088 \pm 0.3084\) & $0\%$*\\
& KAN & 795.2 $\pm$ 257.8 & 0.8579 $\pm$ 0.0155 & 0.7961 $\pm$ 0.0141 & $0\%$*\\
& \textcolor{blue}{dynamic (MLP)} & \textcolor{blue}{\(1,082.72 \pm 229.62\)} & \textcolor{blue}{\textbf{0.0086 $\pm$ 0.0024}} & \textcolor{blue}{\textbf{0.0075 $\pm$ 0.0022}} & \textcolor{blue}{100\%} \\
\hline

\multirow{4}{*}{\textit{Reaction} (Eq.~\ref{eq:reaction})}
& PINNsFormer & 182.2 $\pm$ 34.7 & 0.0266 $\pm$ 0.0110 & 0.0133 $\pm$ 0.0055 & $80\%$ \\
& PINNmamba & 98.3 $\pm$ 35.0 & 0.1052 $\pm$ 0.0464 & 0.0626 $\pm$ 0.0288 & $80\%$ \\
& KAN & 702.9 $\pm$ 152.3 & \textbf{0.0180 $\pm$ 0.0047} & \textbf{0.0076 $\pm$ 0.0020} & 100\% \\
& \textcolor{blue}{dynamic (MLP)} & \textcolor{blue}{\(662.94 \pm 91.11\)} & \textcolor{blue}{\(0.0531 \pm 0.0101\)} & \textcolor{blue}{\(0.0295 \pm 0.0055\)} & \textcolor{blue}{100\%} \\
\hline

\multirow{4}{*}{\textit{Wave} (Eq.~\ref{eq:wave})}
& PINNsFormer &3360.1 $\pm$ 504.3 & 0.3487 $\pm$ 0.0191 & 0.3403 $\pm$ 0.0218 & $0\%$ \\
& PINNmamba & 1540.8 $\pm$ 581.5 & 0.1164 $\pm$ 0.1278 & 0.1134 $\pm$ 0.1256 & $60\%$ \\
& KAN & 3047.8 $\pm$ 700.1 & 0.0776 $\pm$ 0.0262 & 0.0754 $\pm$ 0.0249 & 80\% \\
& \textcolor{blue}{dynamic (MLP)} & \textcolor{blue}{\(1,283.06 \pm 448.48\)} & \textcolor{blue}{\(\mathbf{0.0094} \pm \mathbf{0.0033}\)} & \textcolor{blue}{\(\mathbf{0.0090} \pm \mathbf{0.0030}\)} & \textcolor{blue}{100\%} \\
\hline

\multirow{4}{*}{\textit{Allen--Cahn} (Eq.~\ref{eq:allencahn})}
& PINNsFormer & 3477.7 $\pm$ 260.7 & 0.0666 $\pm$ 0.0134 & 0.0197 $\pm$ 0.0047 & $60\%$ \\
& PINNmamba & 510.2 $\pm$ 187.9 & 0.6787 $\pm$ 0.0897 & 0.4201 $\pm$ 0.0603 & $0\%$*\\
& KAN & 3403.8 $\pm$ 668.7 & 0.0739 $\pm$ 0.0232 & 0.0273 $\pm$ 0.0091 & 100\%\\ 
& \textcolor{blue}{dynamic (MLP)} & \textcolor{blue}{\(3,375.30 \pm 1,140.70\)} & \textcolor{blue}{\(\textbf{0.0495} \pm 0.0097\)} & \textcolor{blue}{\(\textbf{0.0148} \pm 0.0022\)} & \textcolor{blue}{100\%} \\
\bottomrule
\end{tabular}
\end{adjustbox}
\end{center}

A comparison between the vanilla PINN and several advanced architectures proposed to improve convergence is provided in Table~\ref{tab:different_model} and Figure~\ref{fig:benchmark_pred_error}. These architectures can reduce prediction error on some equations when trained purely in FP32. However, their behaviour is not consistently stable across random seeds, especially for the \textit{Convection} and \textit{Allen--Cahn} equations. In contrast, the proposed curvature-aware dynamic precision approach applied to the vanilla MLP achieves robust convergence across all four PDEs. This suggests that even though different architectures can improve PINN convergence for specific problems, they may not be sufficient in all cases and can introduce additional computational cost.

Furthermore, we also test different MLP configurations (Table~\ref{tab:three_layer_mlp} and Figure~\ref{fig:3-layer_speed_vs_error}). With fewer layers and a wider MLP architecture, we observe larger improvements in three out of four benchmark equations. One possible explanation is that architecture depth and width change the optimisation geometry, which may affect the sensitivity of PINN training to numerical precision. Previous studies have shown that network architecture can influence loss-landscape sharpness and curvature, and Hessian-based analyses have observed large isolated eigenvalues during deep-network training~\citep{li2018visualizing,ghorbani2019effect}.

\begin{center}
\begin{minipage}{0.98\textwidth}
    \centering
    \includegraphics[width=0.98\linewidth]{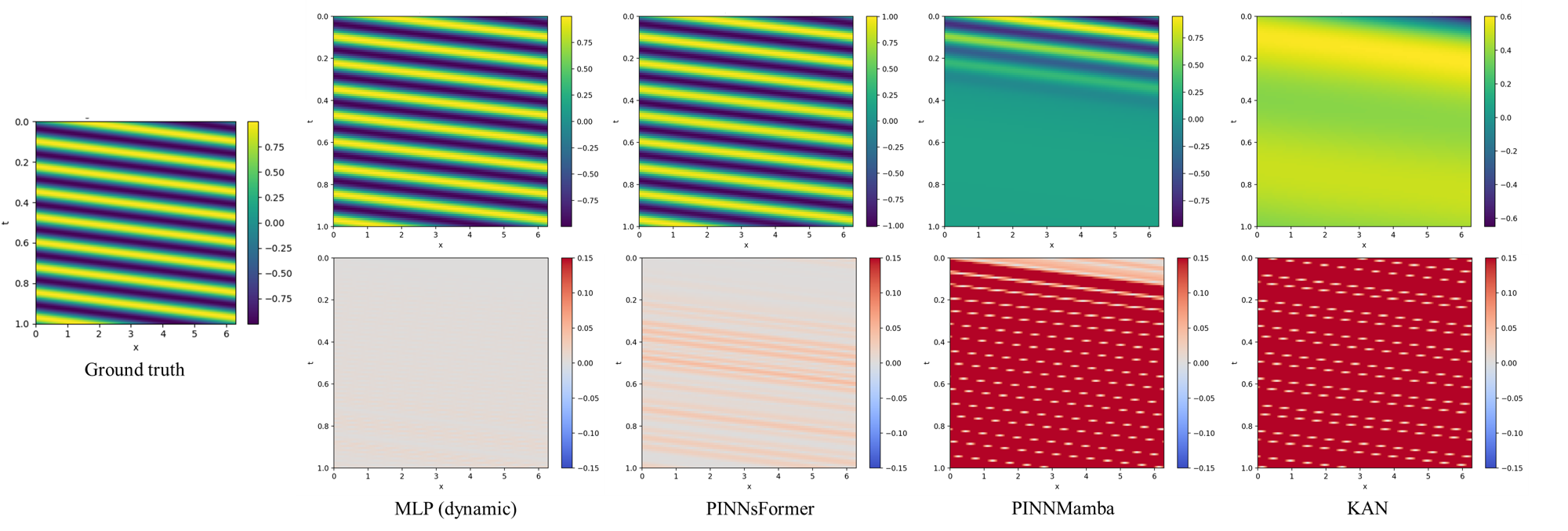}\\[-0.3em]
    {\small (a) \textit{Convection}\par}
    \vspace{0.6em}
    \includegraphics[width=0.98\linewidth]{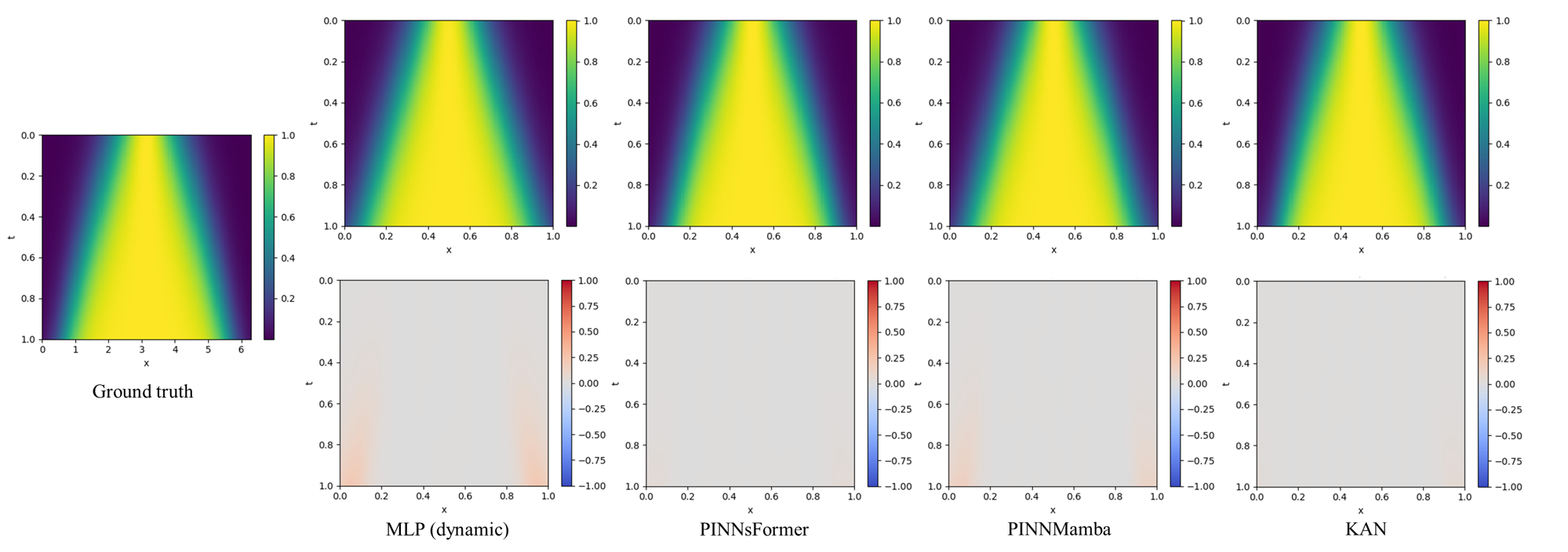}\\[-0.3em]
    {\small (b) \textit{Reaction}\par}
    \captionsetup{hypcap=false}
    \captionof{figure}{Predictions of different network architectures on PINN failure mode benchmark equations (Convection and \textit{Reaction}). Each sub-figure corresponds to one PDE, starting with the ground truth (first left), dynamic-precision MLP, PINNsFormer, PINNMamba, and KAN from left to right. This figure visually compares the predicted solution (top) and absolute error (bottom).}
    \label{fig:benchmark_pred_error}
\end{minipage}
\end{center}

\begin{center}
\begin{minipage}{0.98\textwidth}
    \centering
    \includegraphics[width=0.98\linewidth]{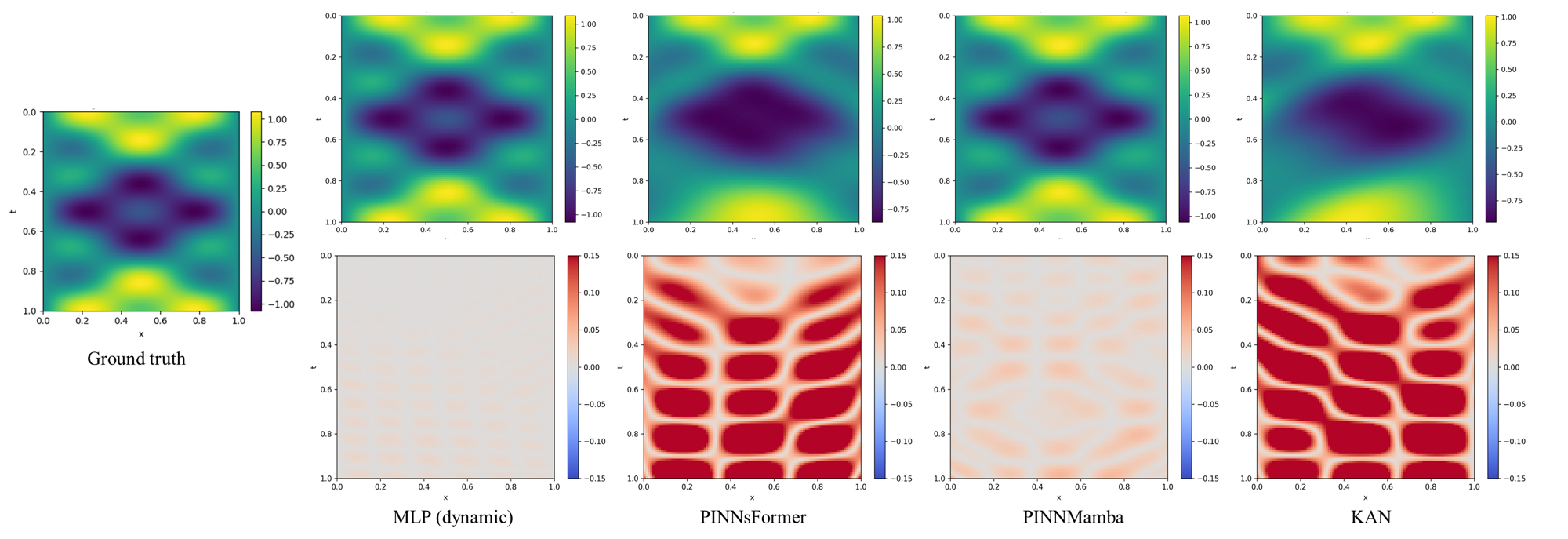}\\[-0.3em]
    {\small (c) \textit{Wave}\par}
    \vspace{0.6em}
    \includegraphics[width=0.98\linewidth]{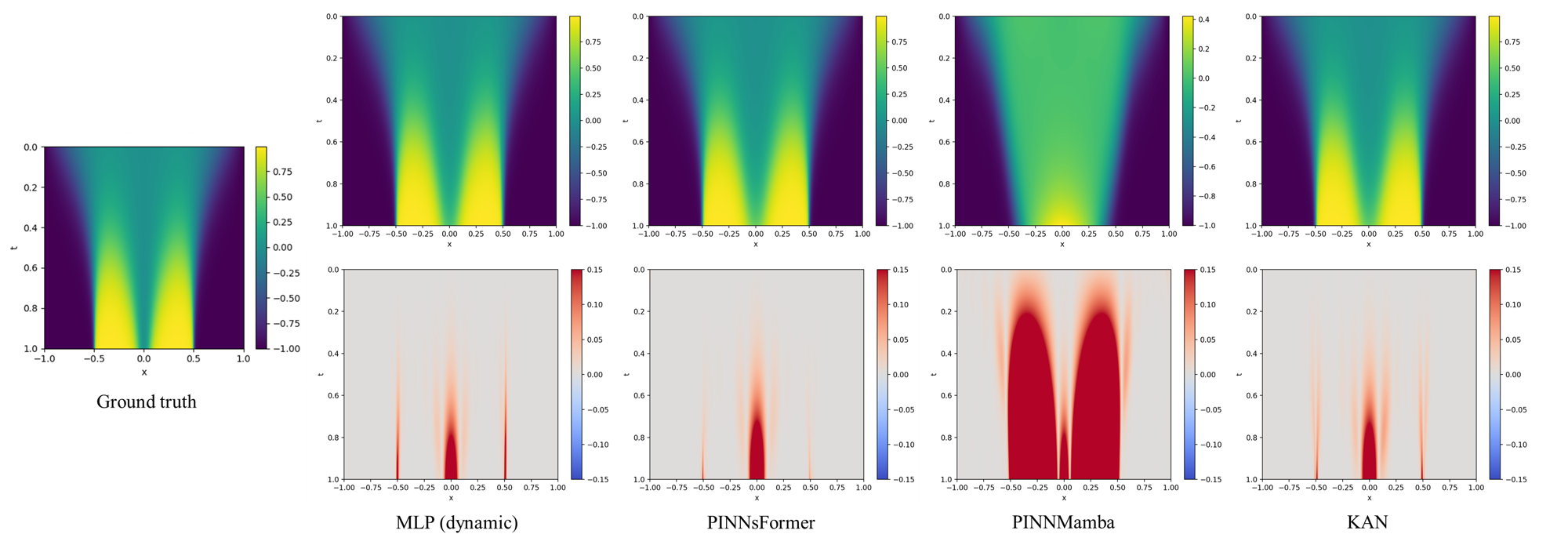}\\[-0.3em]
    {\small (d) Allen--Cahn\par}
    \captionsetup{hypcap=false}
    \captionof*{figure}{Figure~\ref{fig:benchmark_pred_error}, Predictions of different network architectures on PINN failure mode benchmark equations (\textit{Wave} and \textit{Allen--Cahn}). Each sub-figure corresponds to one PDE, starting with the ground truth (first left), dynamic-precision MLP, PINNsFormer, PINNMamba, and KAN from left to right. This figure visually compares the predicted solution (top) and absolute error (bottom).}
    \label{fi}
\end{minipage}
\end{center}

\begin{center}
\begin{minipage}{\linewidth}
    \centering

    \begin{minipage}{0.50\linewidth}
        \centering
        \includegraphics[width=\linewidth]{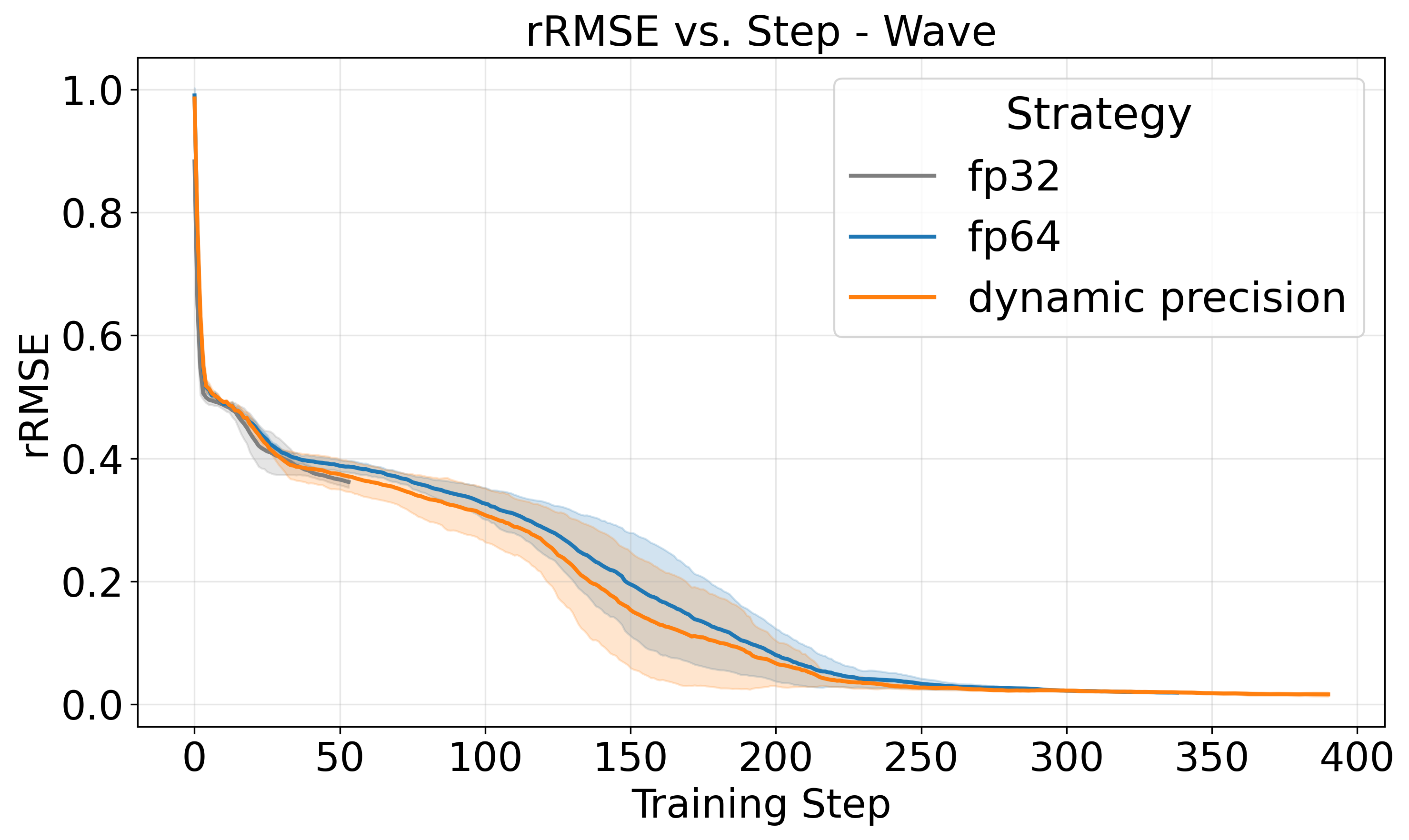}\\[-0.2em]
        {\small (a) Step}
    \end{minipage}%
    \begin{minipage}{0.50\linewidth}
        \centering
        \includegraphics[width=\linewidth]{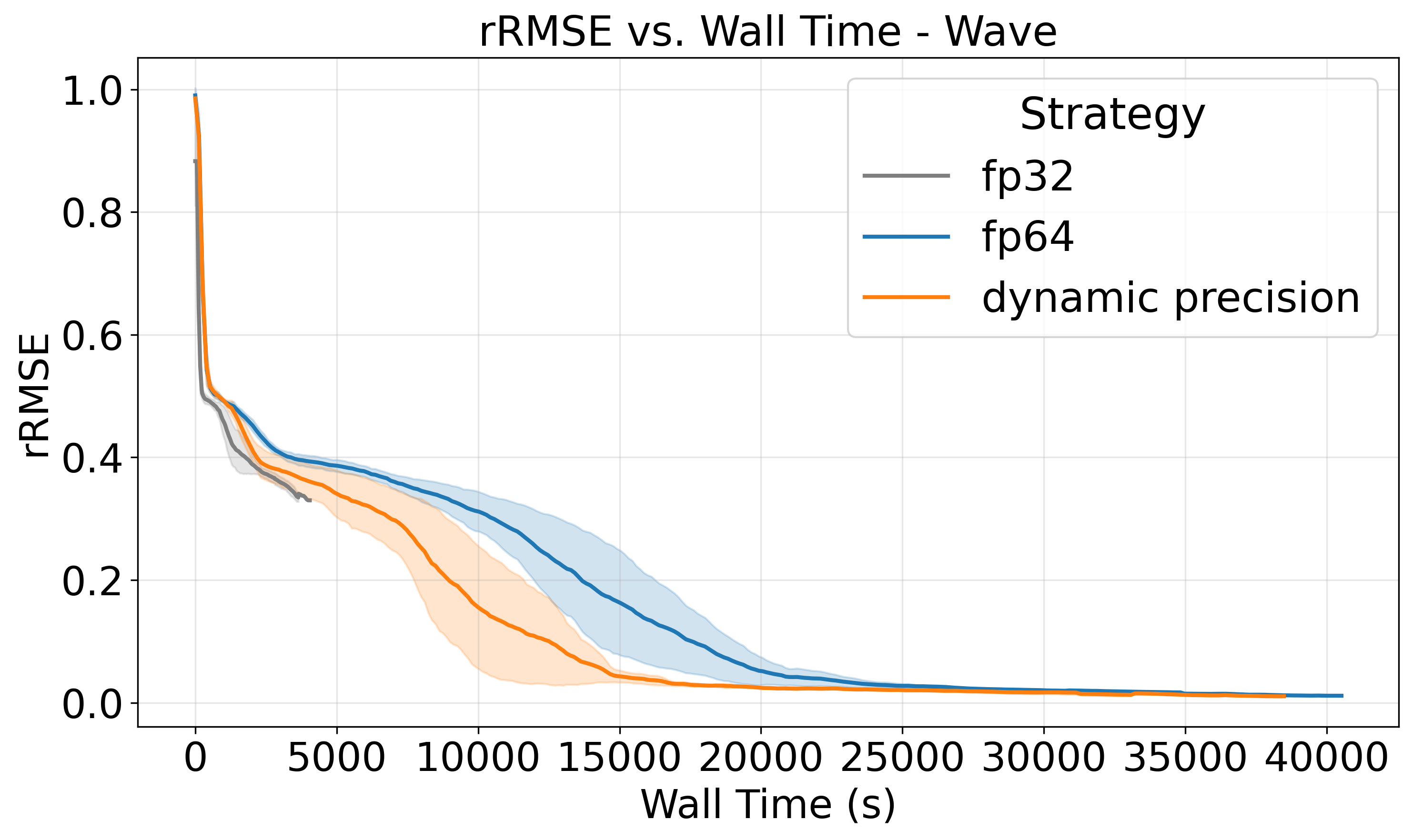}\\[-0.2em]
        {\small (b) Wall time}
    \end{minipage}
    \captionsetup{hypcap=false}
    \captionof{figure}{Step-wise and wall-clock convergence of PINNsFormer on the \textit{Wave} equation. The dynamic approach converges faster both in terms of optimisation steps and wall-clock time.}
    \label{fig:loss_step_vs_loss_time_pinnsformer}
\end{minipage}
\end{center}

\begin{center}
\begin{minipage}{\linewidth}
\centering
\begin{minipage}{0.47\linewidth}
    \centering
    \includegraphics[width=\linewidth]{log_proxy_vs_step_convection_MLP_dynamic.png}\\[-0.25em]
    {\small (a) Vanilla PINN}
\end{minipage}
\hfill
\begin{minipage}{0.47\linewidth}
    \centering
    \includegraphics[width=\linewidth]{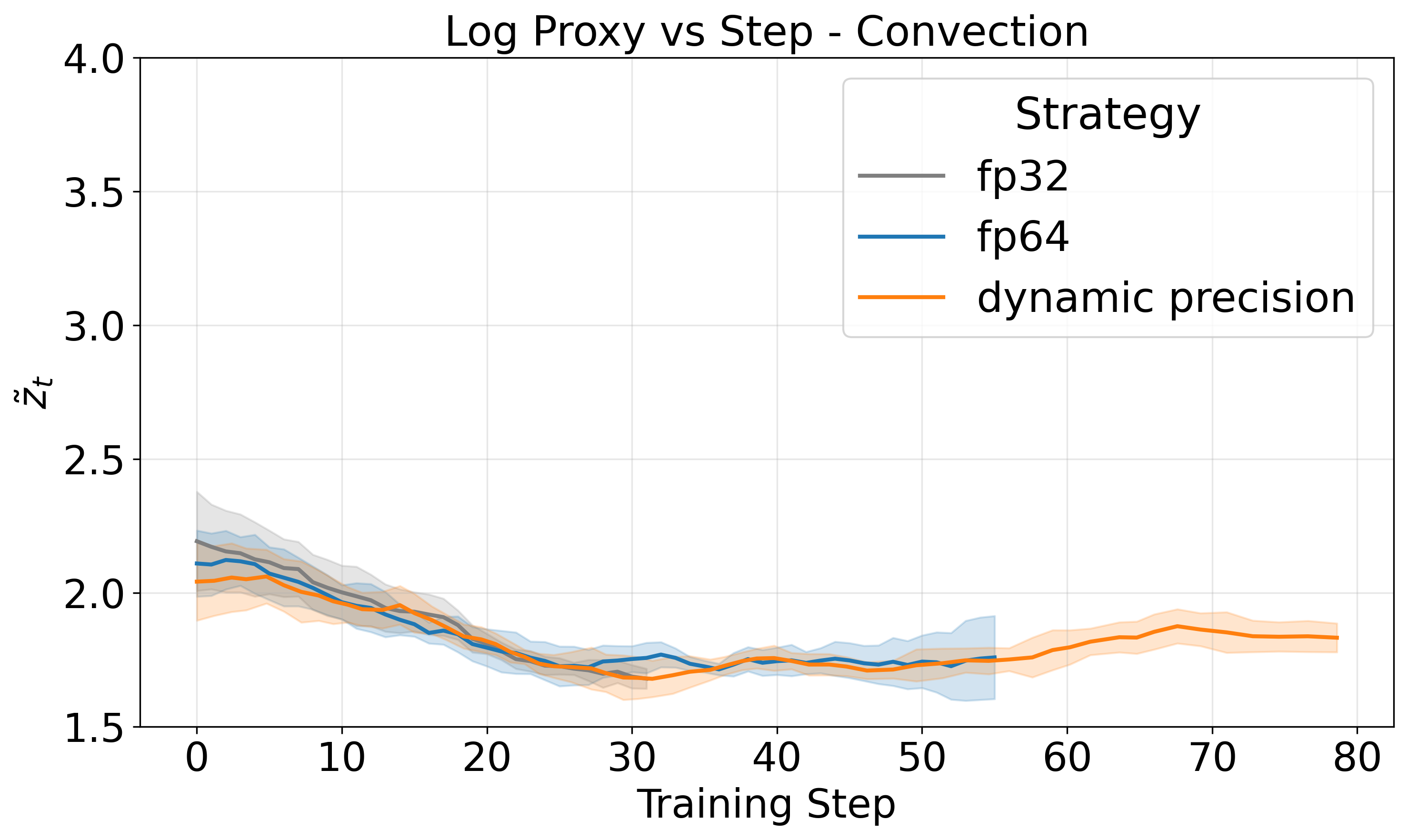}\\[-0.25em]
    {\small (b) PINNsFormer}
\end{minipage}
\captionsetup{hypcap=false}
\captionof{figure}{Curvature comparison between the vanilla PINN and PINNsFormer on the \textit{Convection} benchmark. The plotted signal is the smoothed log-curvature proxy used by the precision controller. Solid lines show the mean over random seeds, and shaded regions indicate the standard deviation. PINNsFormer shows a smoother curvature trajectory on average, suggesting that architecture can affect the conditioning of the precision-switching signal.}\label{fig:curvature_comparision_different_NN}
\end{minipage}
\end{center}

\begin{center}
    \includegraphics[
        width=0.85\textwidth,
        height=0.55\textheight,
        keepaspectratio
    ]{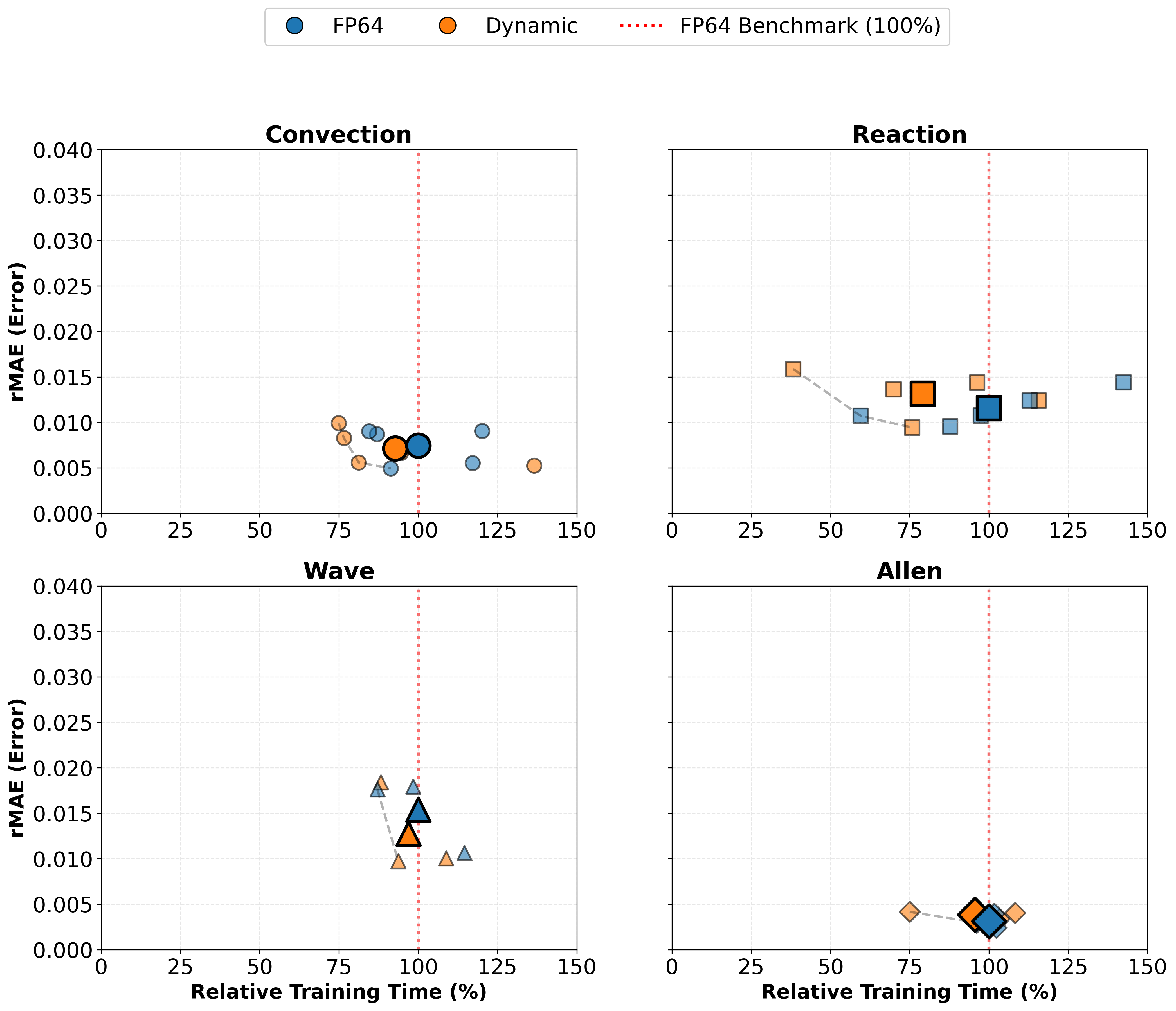}
    \captionsetup{hypcap=false}
    \captionof{figure}{PINNsFormer result for dynamic vs.\ FP64.}
    \label{fig:pinnsformer_dynamic}
\end{center}

To examine the generalisation of the proposed dynamic precision approach, we also apply it to different network architectures (Table~\ref{tab:curvature_aware_PINNsFormer} and Table~\ref{tab:wave_PINNmamba}). The results show that it can also accelerate training with a different architecture, such as PINNsFormer in Figure~\ref{fig:pinnsformer_dynamic}. For the vanilla PINN, the controller mainly improves wall-clock efficiency while showing limited gains in step efficiency (Figures~\ref{fig:rmse_wall_time} and~\ref{fig:rmse_step}), suggesting that its advantage comes mainly from lowering the average cost per iteration by using FP32. In contrast, the PINNsFormer convergence speed in terms of steps is similar to the pure FP64 case, or even better for some equations (Figure~\ref{fig:loss_step_vs_loss_time_pinnsformer}). We hypothesise that the larger improvement when combining curvature-aware dynamic precision with advanced architectures is because these models produce a better-conditioned training trajectory. When comparing the curvature information used as the precision-switching signal (Figure~\ref{fig:curvature_comparision_different_NN}), PINNsFormer shows smoother curvature than the vanilla PINN.
\subsection{Curvature signal and training dynamics}
Rather than being constant throughout optimisation, the numerical sensitivity of PINN training appears to be phase-dependent. Prior work has shown that PINN losses are often difficult to optimise because the differential operators in the residual term induce ill-conditioning in the loss landscape. This makes curvature-related information a plausible signal for adaptive precision control. In our experiments, the curvature proxy \(\tilde{z}_j\) follows a non-monotonic trajectory across different PDEs (Figure~\ref{fig:log_proxy}), indicating that the local optimisation geometry changes over the course of training. This observation supports the use of dynamic rather than fixed precision, since the numerical requirements of training are not uniform across all iterations.
\begin{center}
    \includegraphics[
        width=0.85\textwidth,
        height=0.55\textheight,
        keepaspectratio
    ]{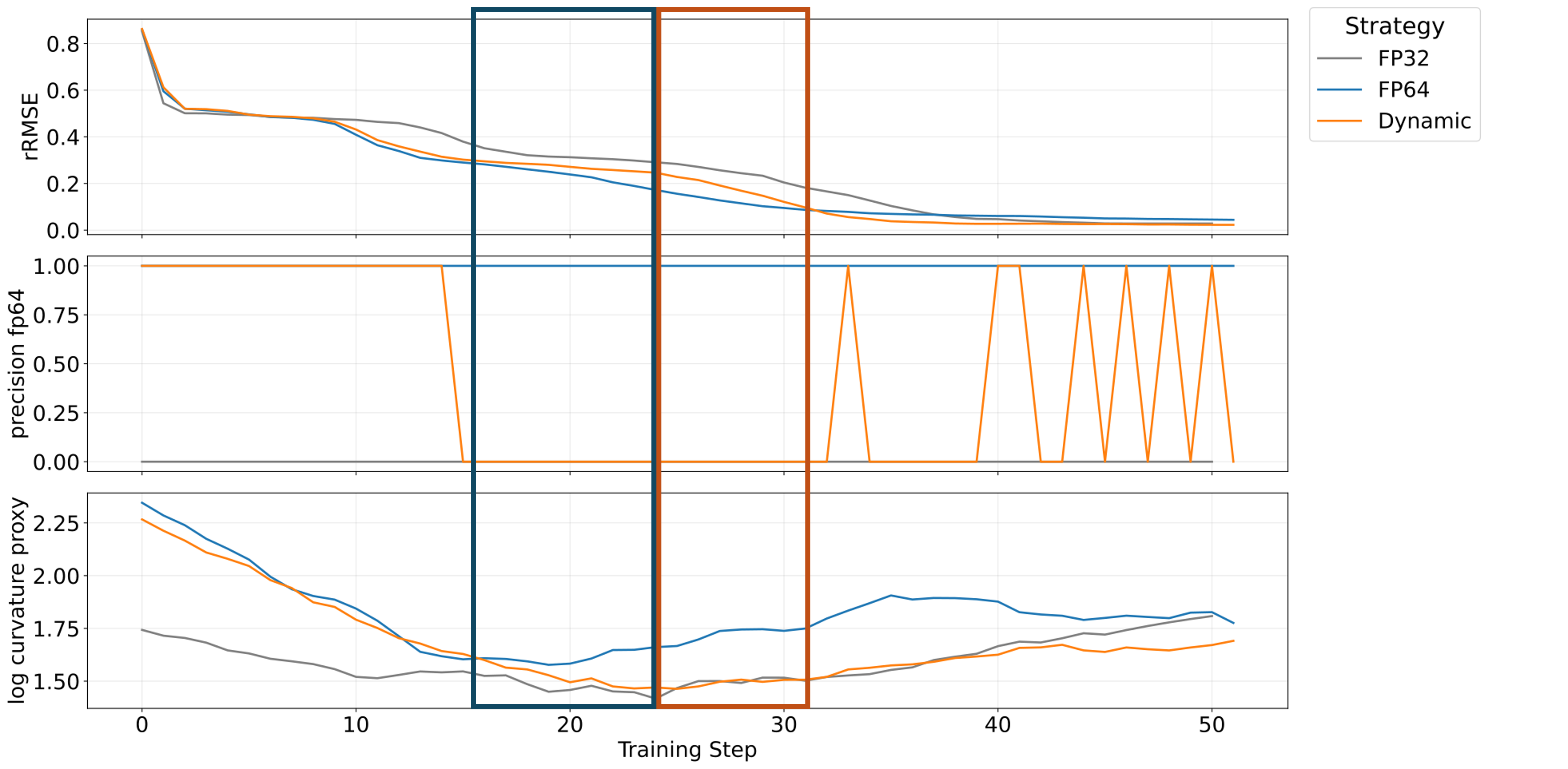}
    \captionsetup{hypcap=false}
    \captionof{figure}{An example of phase-specific precision sensitivity. We show rRMSE (top), precision state (middle), and the smoothed curvature proxy (bottom) during training. Even though the dynamic approach stays in FP32 in both highlighted regions, one phase (blue box) is slower than FP64, while another phase (orange box) is faster. This indicates that convergence speed under FP32 depends on the local curvature regime at different training phases rather than precision alone.}
    \label{fig:example_training_trajectory}
\end{center}

To better understand how the proposed controller affects training, we visualise a representative training trajectory together with its precision state and curvature proxy (Figure~\ref{fig:example_training_trajectory}). The interval in which the dynamic run continues to improve rapidly in FP32 coincides with a relatively low and smooth log-proxy. This supports the intended interpretation of the controller signal: phases with small curvature spread can be handled effectively in lower precision, whereas later increases in the proxy indicate growing numerical sensitivity and motivate promotion back to FP64.

Furthermore, Figure~\ref{fig:example_training_trajectory} shows that FP32 has different effects throughout training. In both highlighted regions, the dynamic approach stays in FP32, but the convergence behaviour is different. In the first highlighted phase, the FP32 trajectory improves more slowly than FP64, while in the second highlighted region, it gives faster wall-clock progress. This indicates that the advantage of lower precision is phase-dependent. When the curvature signal is smooth and stable, the lower per-iteration cost of FP32 can translate into faster wall-clock convergence. When the optimisation becomes more sensitive, the same precision slows down improvement or leads to unstable progress. The proposed controller uses this difference to retain the efficiency benefit of FP32 while recovering the stability of FP64 when needed.

\subsection{Application to an irradiance-driven plant-growth ODE}

PINNs are also gaining attention as prediction tools for real-life dynamical systems. Thus, besides the standard PINN failure model benchmark equations, we also consider a simpler ODE system motivated by irradiance-driven plant growth. For the irradiance ODE equation, the curvature signal is much smoother than the four PINN failure mode benchmark equations (Figure~\ref{fig:irr_speed_vs_accuracy} (a)). This is consistent with the lower-dimensional structure of the ODE problem, which does not include spatial differential operators. However, the system is still nonlinear and depends on the irradiance forcing amplitude, so the optimisation can still be affected by numerical precision.

\begin{center}
\begin{minipage}{0.95\textwidth}
    \centering
    \begin{minipage}[t]{0.48\linewidth}
        \centering
        \includegraphics[width=\linewidth]{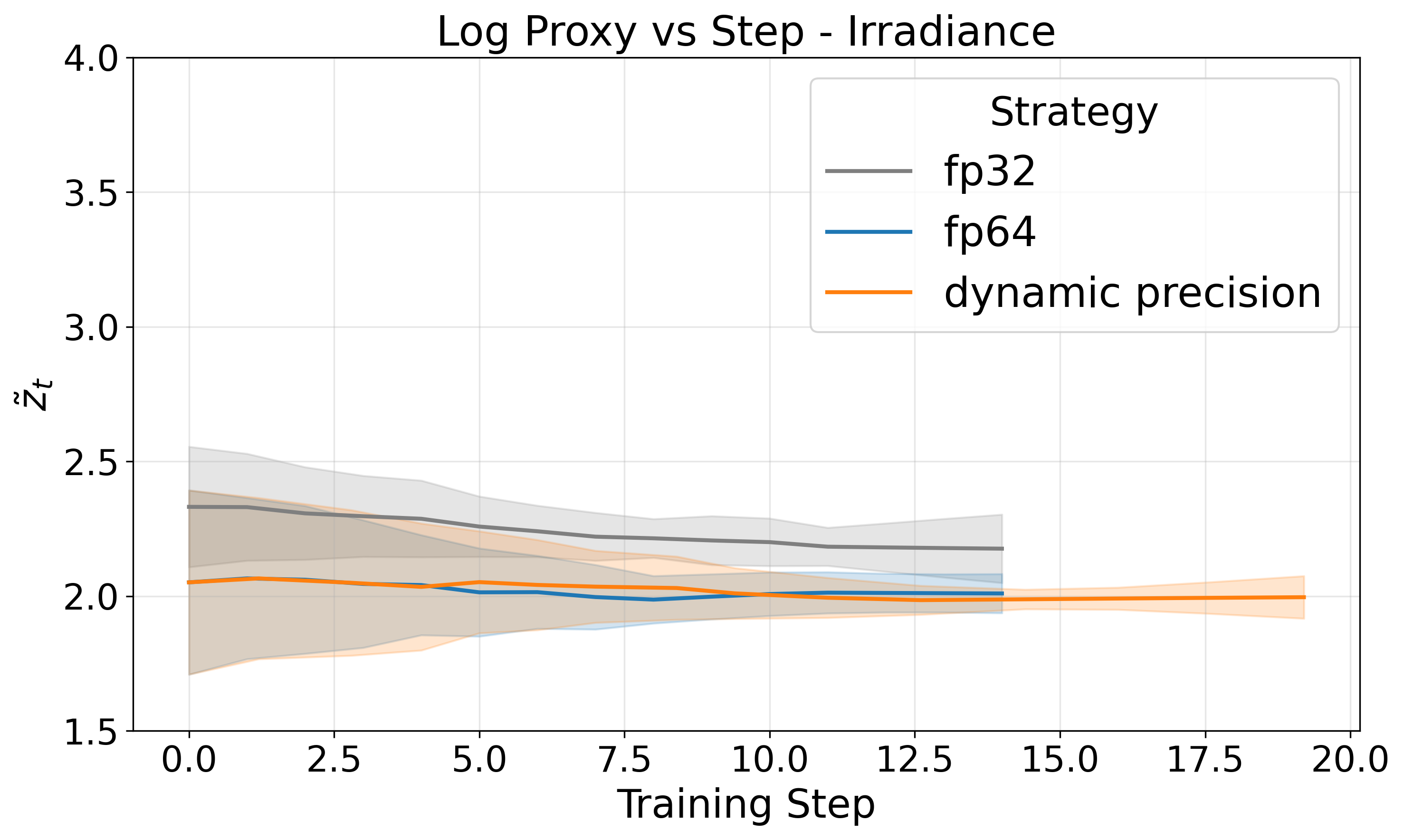}\\[-0.3em]
        {\small (a) Curvature signal\par}
    \end{minipage}
    \hfill
    \begin{minipage}[t]{0.48\linewidth}
        \centering
        \includegraphics[width=\linewidth]{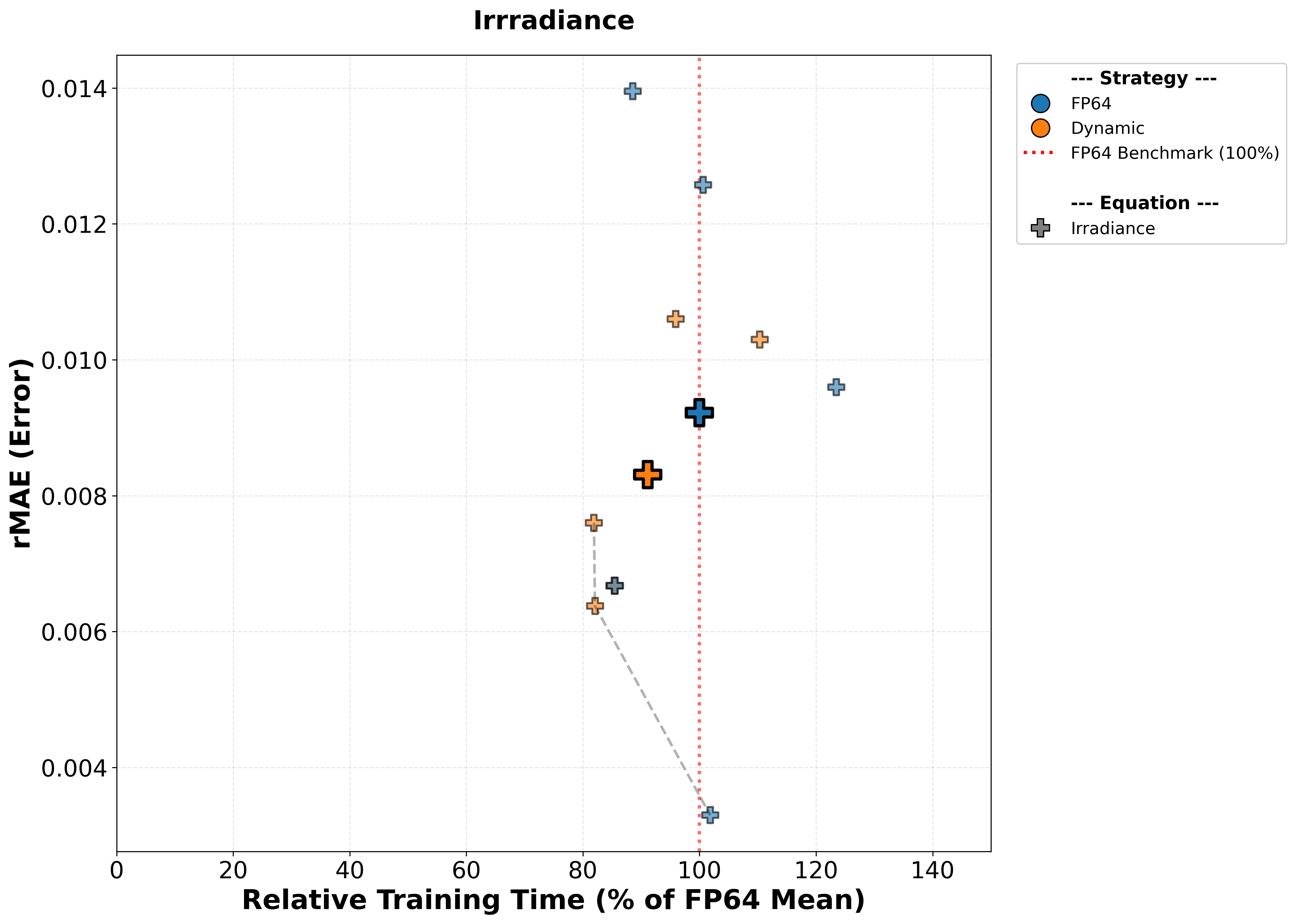}\\[-0.3em]
        {\small (b) Accuracy vs. speed\par}
    \end{minipage}
    \captionsetup{hypcap=false}
    \captionof{figure}{Irradiance ODE curvature signal and result. The proposed dynamic precision approach (orange) also improves both speed and accuracy for the irradiance ODE system.}
    \label{fig:irr_speed_vs_accuracy}
\end{minipage}
\end{center}

The irradiance ODE result shows a similar pattern to the other four benchmark equations with a smoother training trajectory. Dynamic precision reaches an accuracy close to FP64 while reducing the average training time (Figure~\ref{fig:irr_speed_vs_accuracy} (b) and Table~\ref{tab:irradiance_ode}). In comparison, FP32 has a similar runtime but a much larger error variation, indicating that pure single precision can still lead to unstable optimisation for some random seeds even in this smoother ODE setting (Figure~\ref{fig:irradiace_fp32_dynamic} and ~\ref{fig:irradiance_fp32_fp64_dynamic}). Near convergence, small loss and parameter updates can still make the optimiser sensitive to numerical precision. The dynamic approach reduces this risk by switching back to FP64 in precision-sensitive phases; this corresponds to a lower and smoother curvature proxy than fixed FP32 in Figure~\ref{fig:irr_speed_vs_accuracy}(a) and a smaller final error shown in Figure~\ref{fig:irr_speed_vs_accuracy}(b).

\subsection{Limitations and future work}\label{sec:future_work}
Our proposed controller relies on a curvature signal derived from the L-BFGS history. Since L-BFGS is a limited-memory quasi-Newton method, this signal is based on a low-rank approximation constructed from recent secant pairs rather than on the full Hessian~\citep{Nocedal1980UpdatingQM,liuLimitedMemoryBFGS1989}. The signal can be interpreted as a proxy for local curvature along recent update directions, which may not fully capture the local optimisation geometry in highly nonconvex or ill-conditioned regimes. This is particularly relevant for PINNs, whose loss landscapes are known to be challenging to optimise due to ill-conditioning induced by the differential operators~\citep{URBAN2025113656}. In addition, the current controller uses only curvature-related information to decide when to switch precision. In practice, however, reduced-precision stability is also affected by other numerical factors, including gradient magnitude, tensor dynamic range, and underflow or overflow effects~\citep{micikevicius2018mixed,10.5555/3491440.3491844}. These factors are not explicitly captured by our current signal. Different second-order optimisers, such as self-scaled L-BFGS~\citep{URBAN2025113656} and Broyden's quasi-Newton method~\citep{al2014broyden}, have been proposed for PINN training, and future work could adapt the proposed approach to these optimisers employed in PINN training.

In our experiments, specifically designed architectures tend to produce smoother and smaller curvature trajectories(Figure~\ref{fig:curvature_comparision_different_NN} (b)), suggesting that architectural changes can partly improve the conditioning of PINN training. However, these modifications do not completely remove precision-related instability, and their behaviour remains problem-dependent. Furthermore, when using FP64 with those architectures, the training time also increases significantly. Thus, our curvature-aware dynamic precision and architecture design can be viewed as complementary rather than competing approaches for improving PINN training efficiency.

Another challenge is that the curvature signal can be affected by the underlying architecture, as shown in Section~\ref{sec:different_structure}. Even the condition number varies across different PDEs. Thus, we currently treat the switching threshold as a hyperparameter, while it would also be interesting to develop signals that are universal across different structures and equations. A natural direction for future work is therefore to design richer controllers that combine curvature information with additional numerical signals, such as gradient-scale statistics, update norms, or dynamic-range indicators. Another promising direction is to move beyond binary switching between FP32 and FP64 and investigate multi-level adaptive precision strategies. More broadly, it would be valuable to test whether similar precision-control mechanisms are useful in other numerically sensitive scientific machine learning settings where FP64 remains important for stable training and accurate prediction~\citep{kashiMixedprecisionNumericsScientific2026}.

\FloatBarrier
\section{Conclusion}\label{sec:conclusion}
We proposed a curvature-aware dynamic precision approach for PINN training that switches between FP32 and FP64 using curvature information derived from the L-BFGS history. The proposed approach reduces computational cost while maintaining accuracy close to pure FP64 training. Across four PINN failure mode benchmark equations and an ODE example, the method improves the speed--accuracy trade-off compared with fixed FP64 and avoids the instability often observed under fixed FP32. Furthermore, our results suggest that the benefit of dynamic precision is related to training phase-dependent numerical sensitivity during optimisation. We also show that the proposed approach can be applied to different neural network architectures, and that its benefit appears larger when the curvature trajectory is smoother and more stable. Overall, these results indicate that the adaptive precision approach is a practical way to improve the efficiency of PINN training while preserving predictive accuracy, and can be combined with architectural or optimisation-based improvements.
\section{CRediT authorship contribution statement}
\textbf{Yingjie Shao}: Conceptualization; Methodology; Formal Analysis; Visualization; Writing – original draft; Writing – review \& editing
\textbf{Ioannis N. Athanasiadis}: Funding acquisition; Supervision; Writing – review \& editing
\textbf{George van Voorn}: Funding acquisition; Supervision; Writing – review \& editing
\textbf{Taniya Kapoor}: Conceptualization; Supervision; Writing – review \& editing

\section{Data availability}
The Allen--Cahn data were downloaded from the codebase associated with~\citep{xu2025fp64}. The remaining datasets were generated using the scripts and will be provided at our \href{https://github.com/YingjieShao/Curvature_aware_dynamic_precision_for_PINN.git}{GitHub repository} upon publication.

\section{Declaration of competing interest}
The authors declare that they have no known competing financial interests or personal relationships that could have appeared to influence the work reported in this paper.

\section{Acknowledgement}
Thanks to the comments and suggestions given by Fred van Eeuwijk. Fellowship in Data Science and Artificial Intelligence by Wageningen University \& Research.
\newpage

\clearpage
\section*{Appendices}
\addcontentsline{toc}{section}{Appendices}

\begin{appendices}

\renewcommand{\thesection}{\Alph{section}}
\renewcommand{\theequation}{\Alph{section}.\arabic{equation}}
\renewcommand{\thetable}{\Alph{section}.\arabic{table}}
\renewcommand{\thefigure}{\Alph{section}.\arabic{figure}}

\newcommand{\resetappendixcounters}{%
    \setcounter{equation}{0}%
    \setcounter{table}{0}%
    \setcounter{figure}{0}%
}
\section{Notation}
The notations used in this paper are listed in Table~\ref{tab:notation}.
\begin{center}
\centering
\captionsetup{hypcap=false}
\captionof{table}{Summary of notation used in the manuscript.}
\label{tab:notation}
\small
\begin{tabularx}{\linewidth}{@{}p{0.1\linewidth}X p{0.16\linewidth}@{}}
\toprule
\textbf{Symbol} & \textbf{Meaning} & \textbf{Reference} \\
\midrule
\(x\) & Spatial variable & Eq.~\ref{eq:pinn_loss_residual_format} \\
\(\Omega\) & Spatial domain & Eq.~\ref{eq:pinn_loss_residual_format} \\
\(\mathbb{R}^D\) & \(D\)-dimensional Euclidean space & Eq.~\ref{eq:pinn_loss_residual_format} \\
\(D\) & Spatial dimension & Eq.~\ref{eq:pinn_loss_residual_format} \\
\(t\) & Time variable & Eq.~\ref{eq:pinn_loss_residual_format} \\
\(T\) & Final time & Eq.~\ref{eq:pinn_loss_residual_format} \\
\(u(x,t)\), \(u\) & Unknown solution field of the differential equation & Eq.~\ref{eq:pinn_loss_residual_format} \\
\(\mathcal{N}\) & Known differential operator in the forward problem & Eq.~\ref{eq:pinn_loss_residual_format} \\
\(u_\theta(x,t)\) & Neural-network approximation to \(u(x,t)\) & Eq.~\ref{eq:pinn_loss}; Eq.~\ref{eq:pinn_res_loss} \\
\(\theta\) & Trainable neural-network parameters & Eq.~\ref{eq:pinn_loss}; Eq.~\ref{eq:lbfgs_s_y} \\
\midrule
\(\mathcal{L}\) & Total PINN training loss & Eq.~\ref{eq:pinn_loss} \\
\(\mathcal{L}_{\mathrm{res}}\) & PDE residual loss & Eq.~\ref{eq:pinn_loss}; Eq.~\ref{eq:pinn_res_loss} \\
\(\mathcal{L}_B\) & Boundary-condition loss & Eq.~\ref{eq:pinn_loss} \\
\(\mathcal{L}_I\) & Initial-condition loss & Eq.~\ref{eq:pinn_loss} \\
\(\lambda_{\mathrm{res}}\) & Weight of the residual loss & Eq.~\ref{eq:pinn_loss} \\
\(\lambda_B\) & Weight of the boundary-condition loss & Eq.~\ref{eq:pinn_loss} \\
\(\lambda_I\) & Weight of the initial-condition loss & Eq.~\ref{eq:pinn_loss} \\
\(N_{\mathrm{res}}\) & Number of residual collocation points & Eq.~\ref{eq:pinn_res_loss} \\
\((x_i^{\mathrm{res}},t_i^{\mathrm{res}})\) & The \(i\)-th residual collocation point & Eq.~\ref{eq:pinn_res_loss} \\
\(i\) & Index of collocation or evaluation points & Eq.~\ref{eq:pinn_res_loss}; Eq.~\ref{eq:rmse_rmae} \\
\midrule
\(k\) & L-BFGS iteration index used for stored curvature pairs & Eq.~\ref{eq:lbfgs_s_y} \\
\(\theta_k\) & Network parameters at L-BFGS iteration \(k\) & Eq.~\ref{eq:lbfgs_s_y} \\
\(\nabla_\theta \mathcal{L}(\theta_k)\) & Gradient of the loss with respect to \(\theta\) at iteration \(k\) & Eq.~\ref{eq:lbfgs_s_y} \\
\(s_k\) & Parameter-update difference, \(s_k=\theta_{k+1}-\theta_k\) & Eq.~\ref{eq:lbfgs_s_y} \\
\(y_k\) & Gradient-update difference, \(y_k=\nabla_\theta\mathcal{L}(\theta_{k+1})-\nabla_\theta\mathcal{L}(\theta_k)\) & Eq.~\ref{eq:lbfgs_s_y} \\
\(B_{k+1}\) & Quasi-Newton approximation of the Hessian & Eq.~\ref{eq:kappa_k} \\
\(\kappa_k\) & Secant-based directional curvature estimate along \(s_k\) & Eq.~\ref{eq:kappa_k} \\
\midrule

\(j\) & Outer L-BFGS optimisation step used by the controller & Eq.~\ref{eq:consition_number_proxy}; Algorithm~\ref{alg:curvature_controller} \\
\(\mathcal{H}_j\) & Set of valid stored L-BFGS curvature pairs at step \(j\) & Eq.~\ref{eq:consition_number_proxy}; Algorithm~\ref{alg:curvature_controller} \\
\(\hat{\kappa}_{\mathrm{cond}}\) & Limited-memory conditioning proxy computed from valid curvature pairs & Eq.~\ref{eq:consition_number_proxy}; Eq.~\ref{eq:smooth_curvature} \\
\(z_j\) & Log-transformed conditioning proxy at step \(j\) & Eq.~\ref{eq:smooth_curvature} \\
\(\tilde{z}_j\) & Exponentially smoothed log-conditioning proxy & Eq.~\ref{eq:smooth_curvature}; Eq.~\ref{eq:condition_slope} \\
\(\Delta\tilde{z}_j\) & Slope/change of the smoothed conditioning proxy & Eq.~\ref{eq:condition_slope} \\
\(\varepsilon\) & Small numerical-stability constant used in the log transform, set to \(10^{-12}\) & Eq.~\ref{eq:smooth_curvature} \\
\(\alpha\) & Exponential moving average smoothing coefficient, set to \(0.9\) & Eq.~\ref{eq:smooth_curvature} \\
\(\tau_z\) & Threshold applied to \(\tilde{z}_j\) for precision switching & Algorithm~\ref{alg:curvature_controller} \\
\(J_{\max}\) & Maximum number of L-BFGS optimisation steps & Algorithm~\ref{alg:curvature_controller} \\
\midrule

\(N\) & Number of evaluation points on the benchmark grid & Eq.~\ref{eq:rmse_rmae} \\
\(u(x_i,t_i)\) & Reference solution value at evaluation point \((x_i,t_i)\) & Eq.~\ref{eq:rmse_rmae} \\
\(u_\theta(x_i,t_i)\) & PINN prediction at evaluation point \((x_i,t_i)\) & Eq.~\ref{eq:rmse_rmae} \\

\midrule

\(A\) & Amplitude of the simulated seasonal irradiance driver, and we set \(A\in[-1,1]\) & Eq.~\ref{eq:irradiance} \\
\(S(t;A)\) & Simulated irradiance effect on biomass growth rate & Eq.~\ref{eq:irradiance} \\
\(r\) & Baseline growth-rate parameter in the irradiance ODE & Eq.~\ref{eq:irradiance} \\
\(K\) & Carrying-capacity parameter in the logistic growth term & Eq.~\ref{eq:irradiance} \\
\(u(t;A)\) & Biomass solution of the irradiance-driven ODE for amplitude \(A\) in time interval \(t\in[0,1]\) & Eq.~\ref{eq:irradiance} \\
\(u_\theta(t,A)\) & PINN approximation to the irradiance-driven solution family & Eq.~\ref{eq:irradiance} \\
\bottomrule
\end{tabularx}
\end{center}
\section{Ablation study of threshold}
\resetappendixcounters
We treat the curvature signal \(\tilde{z}_j\) as a hyperparameter and tune for each equation. Table~\ref{tab:different_threshold_ablation} shows the result for different equations under different precision switching thresholds. Across the tested thresholds, the dynamic approach generally maintains FP64-level accuracy while reducing or matching FP64 training time.
\vspace{-0.3em}
\begin{center}
\centering
\captionsetup{hypcap=false}
\captionof{table}{Average results over 5 random seeds for the dynamic precision controller under different \(\tilde{z}_j\) thresholds (with vanilla PINN).}
\label{tab:different_threshold_ablation}
\begin{adjustbox}{max width=\textwidth}
\begin{tabular}{llcccc}
\toprule
Threshold & Metric & \textit{Convection} & \textit{Reaction} & \textit{Wave} & \textit{Allen--Cahn}\\
\midrule
\multirow{3}{*}{2.0}
& Time (s)   & \(\mathbf{1,082.72 \pm 229.62}\) & \(663.16 \pm 107.89\) & \(1,320.94 \pm 499.42\) & \(3,715.00 \pm 923.54\) \\
& rRMSE      & \(0.0086 \pm 0.0024\) & \(0.0545 \pm 0.0142\) & \(0.0112 \pm 0.0057\) & \(0.0503 \pm 0.0067\) \\
& rMAE       & \(0.0075 \pm 0.0022\) & \(0.0304 \pm 0.0082\) & \(0.0109 \pm 0.0052\) & \(0.0147 \pm 0.0018\) \\
\midrule
\multirow{3}{*}{2.5}
& Time (s)   & \(1,089.00 \pm 242.26\) & \(\mathbf{662.94 \pm 91.11}\) & \(\mathbf{1,283.06 \pm 448.48}\) & \(3,658.57 \pm 870.73\) \\
& rRMSE      & \(\mathbf{0.0063 \pm 0.0012}\) & \(\mathbf{0.0531 \pm 0.0101}\) & \(\mathbf{0.0094 \pm 0.0033}\) & \(\mathbf{0.0495 \pm 0.0068}\) \\
& rMAE       & \(\mathbf{0.0055 \pm 0.0010}\) & \(\mathbf{0.0295 \pm 0.0055}\) & \(\mathbf{0.0090 \pm 0.0030}\) & \(\mathbf{0.0145 \pm 0.0019}\) \\
\midrule
\multirow{3}{*}{3.0}
& Time (s)   & -- & -- & -- & \(\mathbf{3,375.30 \pm 1,140.70}\) \\
& rRMSE      & -- & -- & -- & \(\mathbf{0.0495 \pm 0.0097}\) \\
& rMAE       & -- & -- & -- & \(0.0148 \pm 0.0022\) \\
\bottomrule
\end{tabular}
\end{adjustbox}
\begin{tablenotes}
\footnotesize
\item Speedup is computed relative to the FP64 baseline for the corresponding benchmark. Best results across threshold values for each equation are shown in bold. We only ran threshold \(\,=3\) for \textit{Allen--Cahn} because, for the other PDEs, \(\tilde{z}_j\) always remains below 3, as shown in Figure~\ref{fig:log_proxy}.
\end{tablenotes}
\end{center}
\vspace{-1.0em}
\section{Example of precision switching}
\resetappendixcounters
We plot an example to show the precision used during training for our curvature-aware dynamic approach (Figure~\ref{fig:precision_switching_example}).
\begin{center}
\begin{minipage}{\linewidth}
\centering
\newcommand{\switchpanelheight}{0.18\textheight}

\begin{minipage}{0.50\linewidth}
    \centering
    \includegraphics[height=\switchpanelheight]{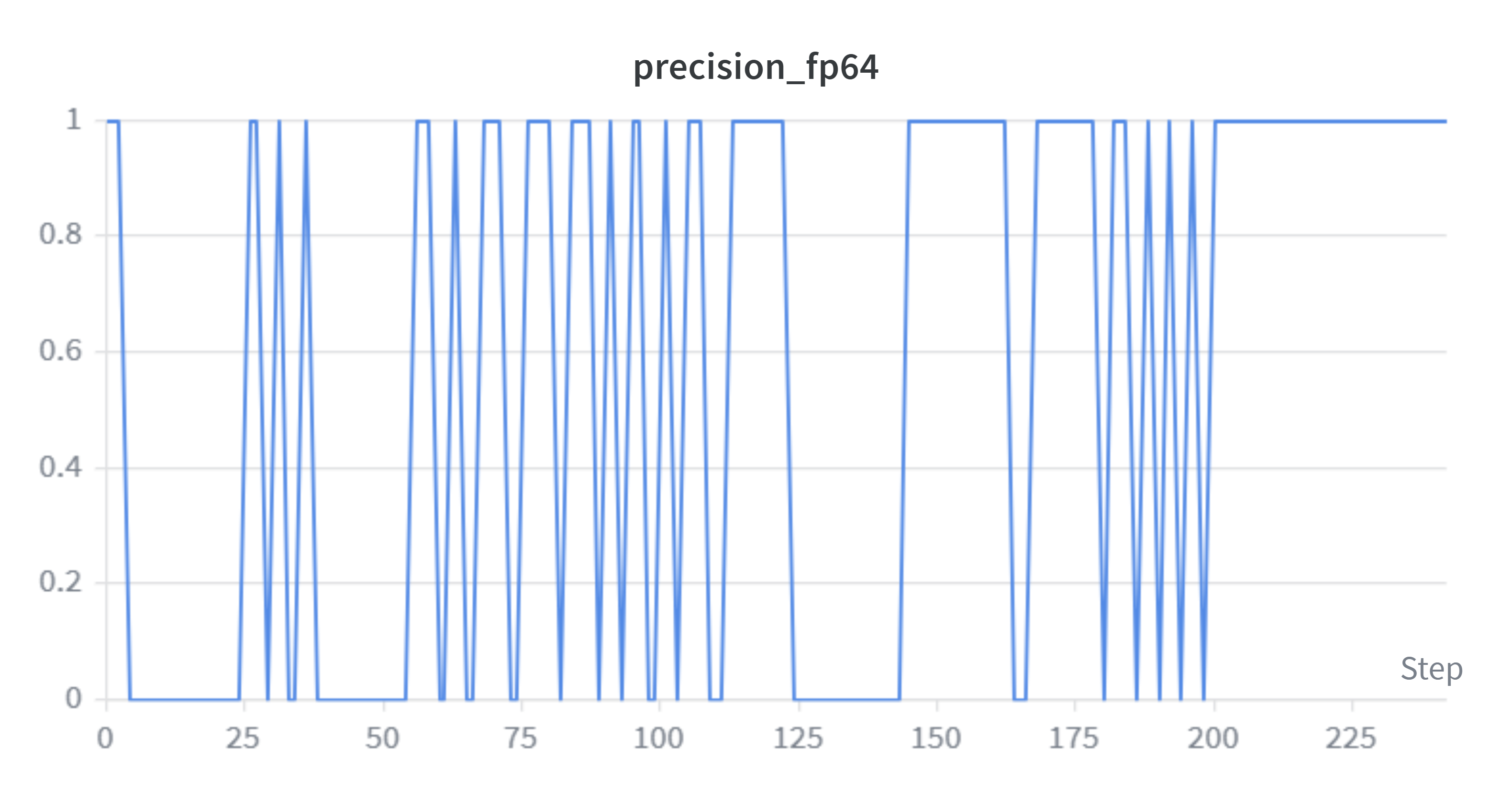}\\[-0.35em]
    {\small (a) \textit{Convection}}
\end{minipage}%
\begin{minipage}{0.50\linewidth}
    \centering
    \includegraphics[height=\switchpanelheight]{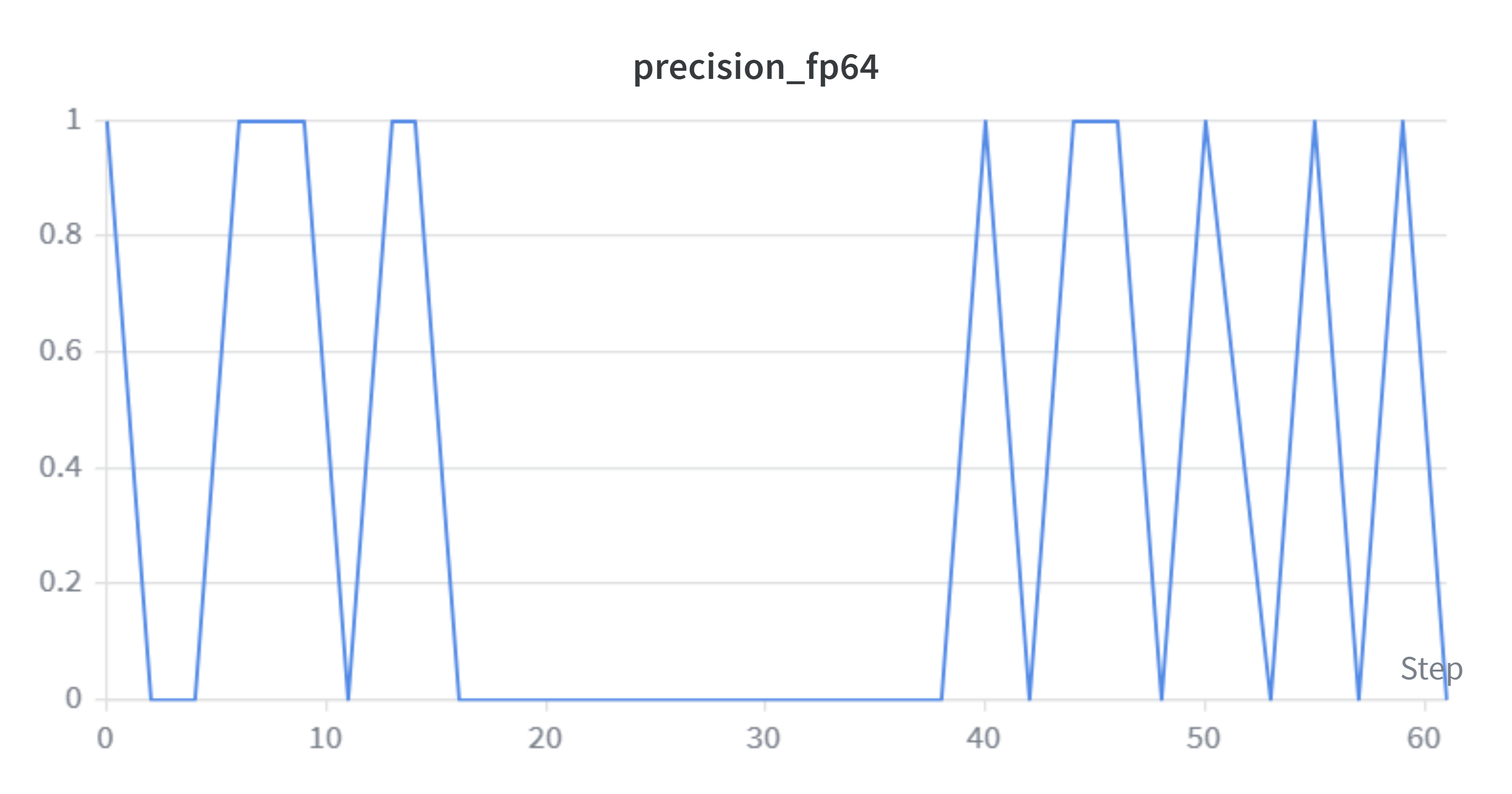}\\[-0.35em]
    {\small (b) \textit{Reaction}}
\end{minipage}\\[-0.2em]
\begin{minipage}{0.50\linewidth}
    \centering
    \includegraphics[height=\switchpanelheight]{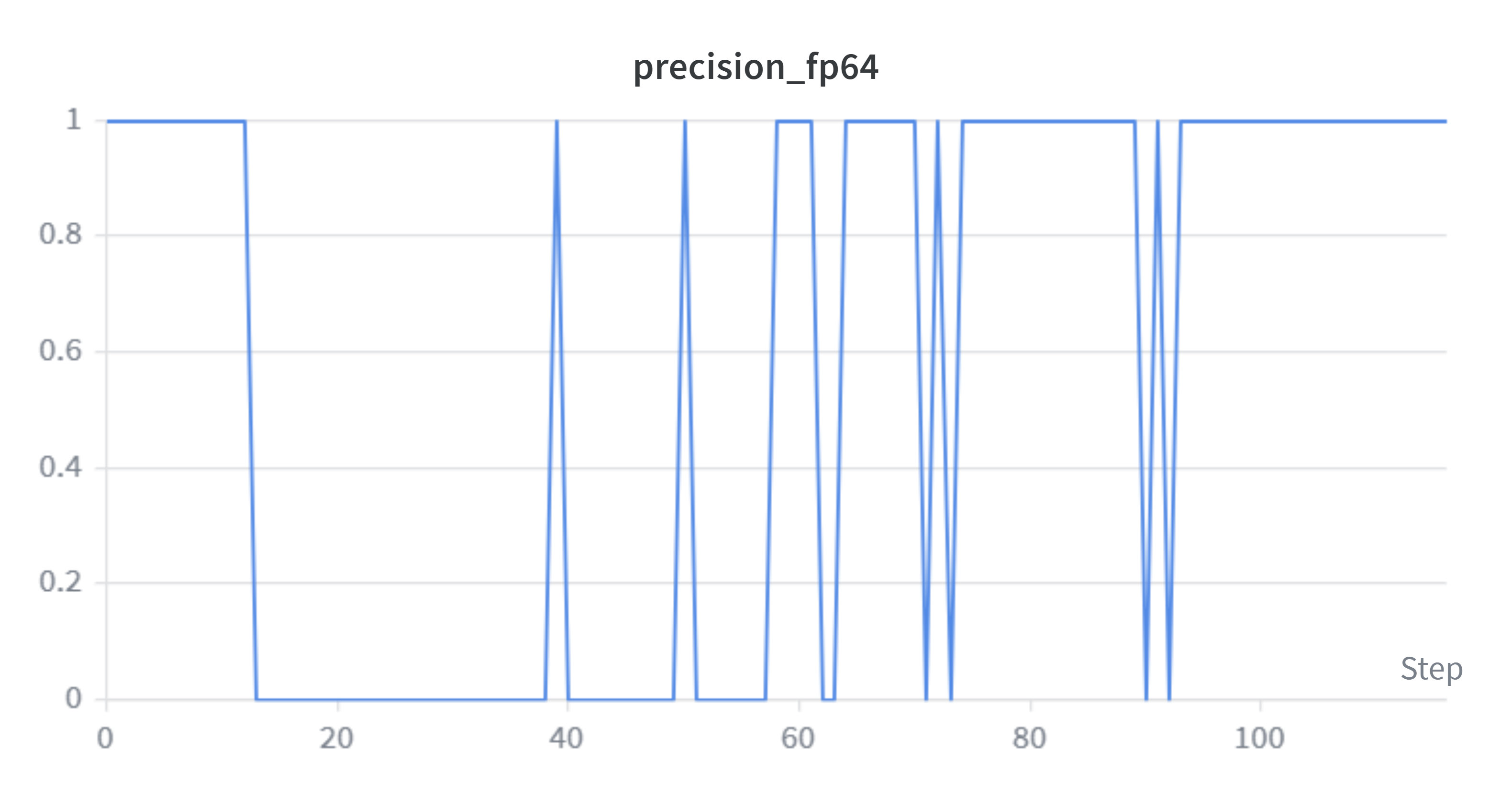}\\[-0.35em]
    {\small (c) \textit{Wave}}
\end{minipage}%
\begin{minipage}{0.50\linewidth}
    \centering
    \includegraphics[height=\switchpanelheight]{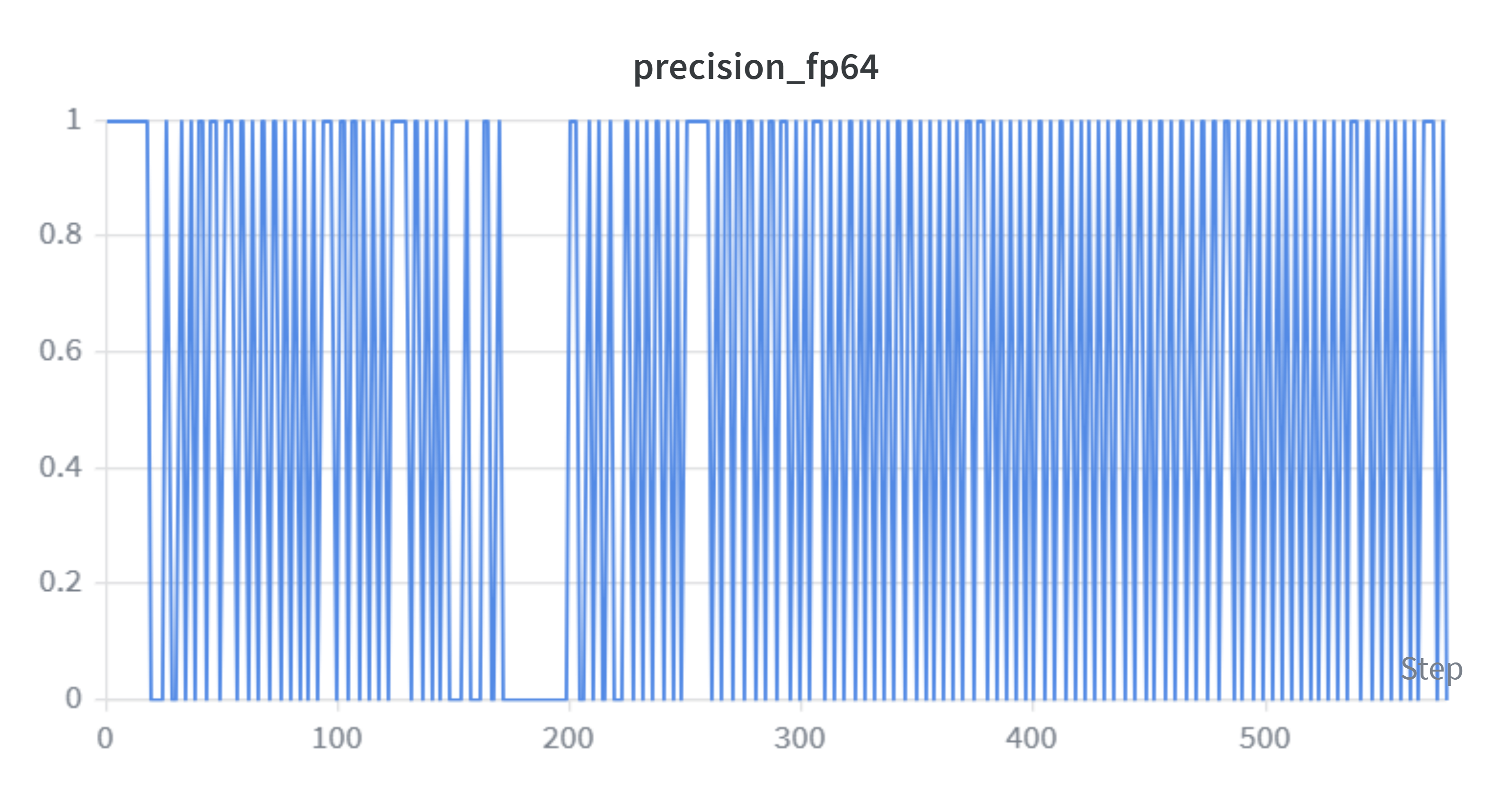}\\[-0.35em]
    {\small (d) Allen--Cahn}
\end{minipage}
\captionsetup{hypcap=false}
\captionof{figure}{Precision switching during training examples from four PDEs. The y-axis indicates the active precision state, where 1 denotes FP64 and 0 denotes FP32. In general, the model stays in FP64 longer and switches more often during training for the \textit{Convection} and \textit{Allen--Cahn} equations, consistent with their more fluctuating and higher curvature signals.}
\label{fig:precision_switching_example}
\end{minipage}
\end{center}

\section{MLP Loss curve during training}
\resetappendixcounters
We plot the prediction error (rRMSE) against wall-clock time (Figure~\ref{fig:rmse_wall_time}) and optimisation steps (Figure~\ref{fig:rmse_step}) to show that, for the vanilla PINN architecture, the wall-clock-time improvement mainly comes from a shorter time per optimisation step rather than faster step-wise convergence.
\begin{center}
\centering
\includegraphics[width=0.9\linewidth]{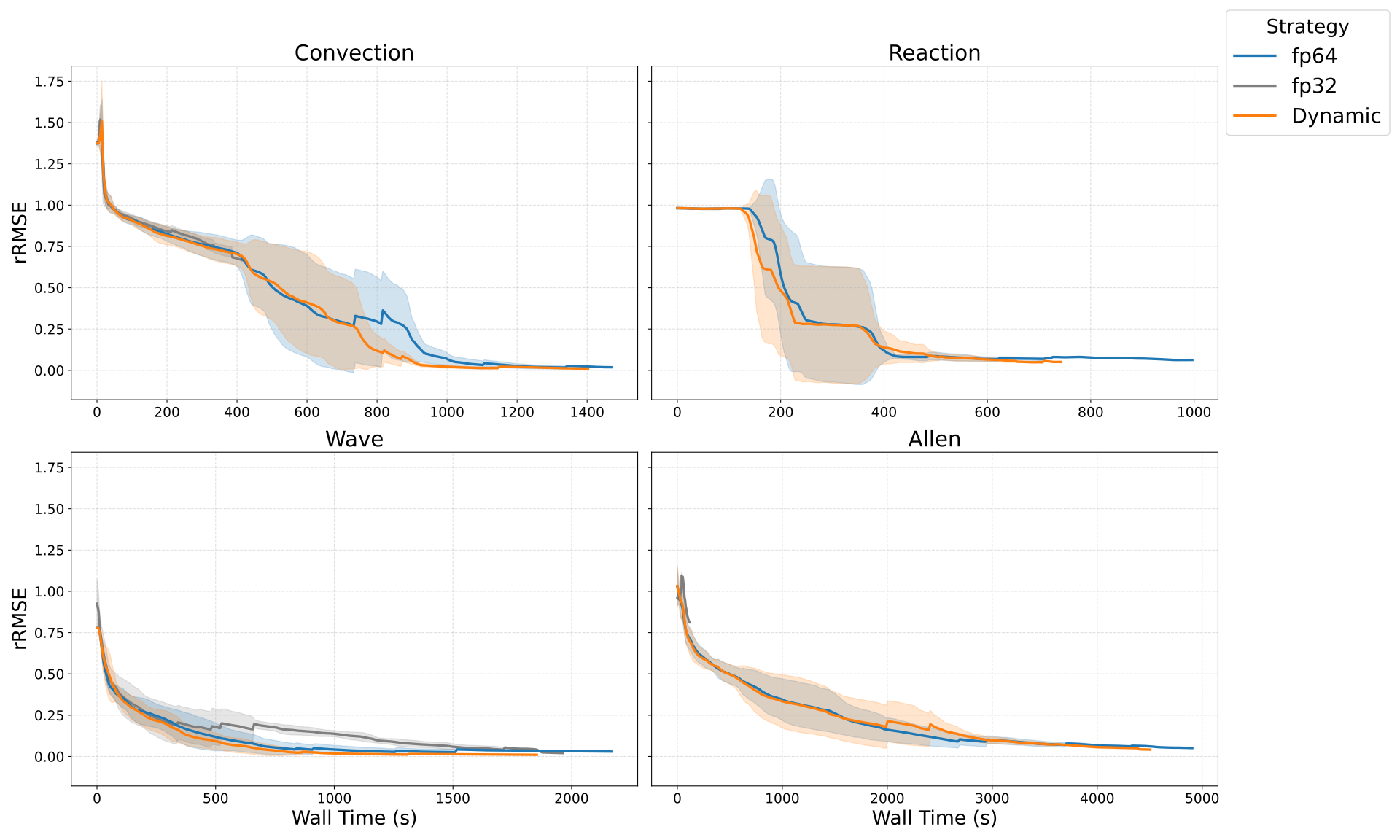}
\captionsetup{hypcap=false}
\captionof{figure}{rRMSE against wall-clock time, evaluated on a common wall-time grid by interpolation. The dynamic approach converges faster in wall-clock time compared with FP64.}
\label{fig:rmse_wall_time}
\end{center}

\begin{center}
\centering
\includegraphics[width=0.9\linewidth]{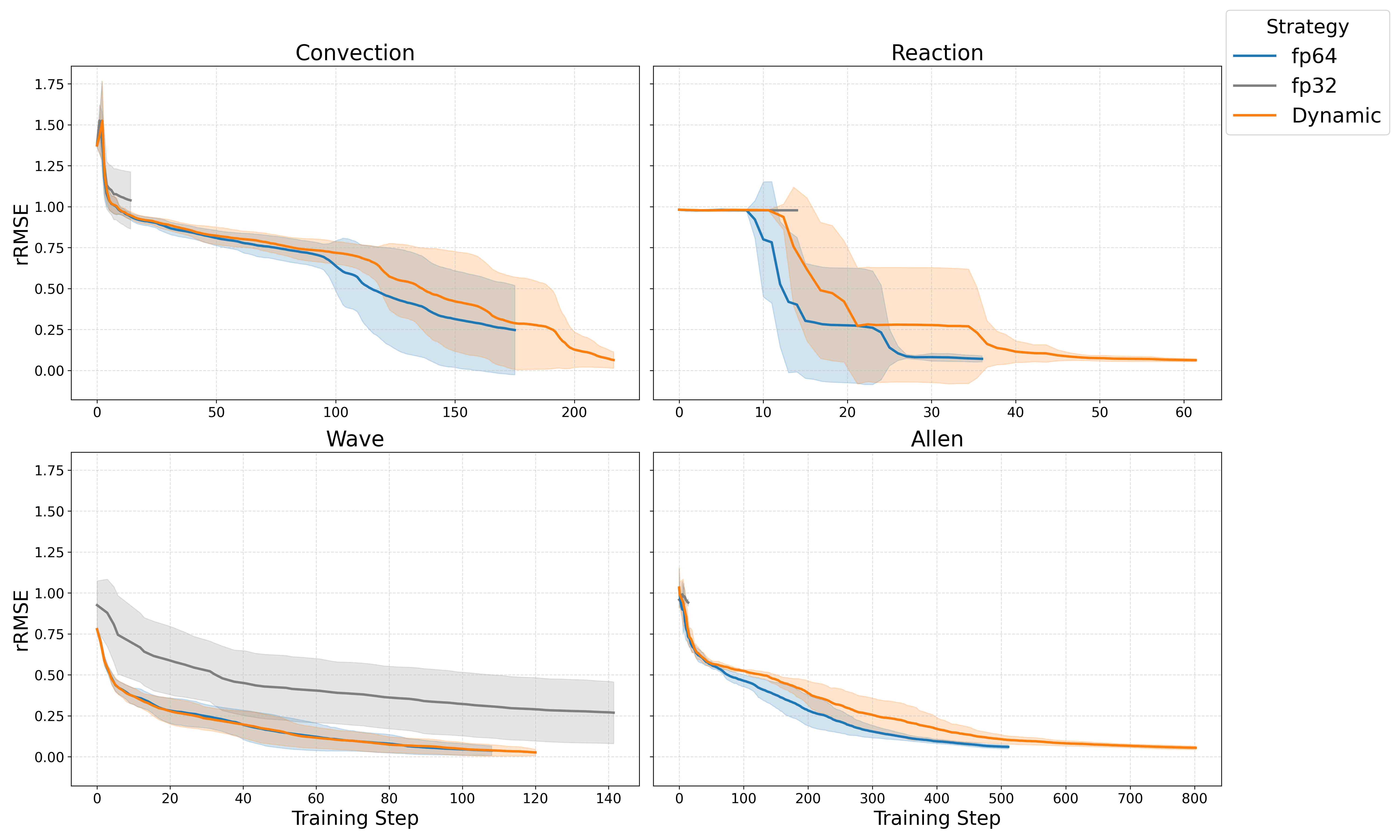}
\captionsetup{hypcap=false}
\captionof{figure}{rRMSE against optimisation steps for the four benchmark PDEs. FP64 generally converges faster in terms of optimisation steps for the vanilla PINN.}
\label{fig:rmse_step}
\end{center}

\newpage
\section{Three-layer Vanilla PINN result}
\resetappendixcounters
To examine whether the proposed controller can generalise to different model sizes, we use the same setup as vanilla PINN but reduce the number of hidden layers while increasing the hidden dimension (3 hidden layers and 648 hidden dimensions).
\begin{center}
\captionsetup{hypcap=false}
\captionof{table}{Three-layer Vanilla PINN average result from 5 random seeds.}
\label{tab:three_layer_mlp}
\begin{adjustbox}{max width=\textwidth}
\begin{tabular}{cccccc}
\toprule
Equation & approach & Avg Time (s) & rRMSE & rMAE & Speedup vs FP64 \\
\midrule
\textcolor{blue}{\textit{Convection}} & \textcolor{blue}{dynamic} & \textcolor{blue}{\(\mathbf{1,679.02} \pm \mathbf{215.59}\)} & \textcolor{blue}{\(\mathbf{0.0233} \pm \mathbf{0.0177}\)} & \textcolor{blue}{\(\mathbf{0.0203} \pm \mathbf{0.0161}\)} & \textcolor{blue}{\(\mathbf{1.12}\times\)} \\
Convection & FP64 & \(1,877.32 \pm 184.31\) & \(0.0281 \pm 0.0317\) & \(0.0256 \pm 0.0306\) & \(1.00\times\) \\
\hline
\textcolor{blue}{\textit{Reaction}} & \textcolor{blue}{dynamic} & \textcolor{blue}{\(\mathbf{169.24} \pm \mathbf{36.01}\)} & \textcolor{blue}{\(0.0594 \pm 0.0072\)} & \textcolor{blue}{\(0.0345 \pm 0.0051\)} & \textcolor{blue}{\(\mathbf{1.11}\times\)} \\
\textit{Reaction} & FP64 & \(187.13 \pm 68.97\) & \(\textbf{0.0559} \pm 0.0233\) & \(\textbf{0.0321} \pm 0.0142\) & \(1.00\times\) \\
\hline
\textcolor{blue}{\textit{Wave}} & \textcolor{blue}{dynamic} & \textcolor{blue}{\(\mathbf{952.05} \pm \mathbf{236.34}\)} & \textcolor{blue}{\(\mathbf{0.0084} \pm \mathbf{0.0034}\)} & \textcolor{blue}{\(\mathbf{0.0082} \pm \mathbf{0.0031}\)} & \textcolor{blue}{\(\mathbf{1.28}\times\)} \\
\textit{Wave }& FP64 & \(1,214.00 \pm 367.24\) & \(0.0101 \pm 0.0029\) & \(0.0101 \pm 0.0030\) & \(1.00\times\) \\
\hline
\textcolor{blue}{\textit{Allen}} & \textcolor{blue}{dynamic} & \textcolor{blue}{\(\mathbf{5,983.74} \pm \mathbf{1,165.95}\)} & \textcolor{blue}{\(\mathbf{0.0427} \pm \mathbf{0.0079}\)} & \textcolor{blue}{\(\mathbf{0.0120} \pm \mathbf{0.0024}\)} & \textcolor{blue}{\(\mathbf{1.01}\times\)} \\
\textit{Allen} & FP64 & \(6,032.66 \pm 1,105.16\) & \(0.0435 \pm 0.0095\) & \(0.0124 \pm 0.0028\) & \(1.00\times\) \\
\bottomrule
\end{tabular}
\end{adjustbox}
\end{center}

\begin{center}
    \centering
    \includegraphics[width=0.97\linewidth]{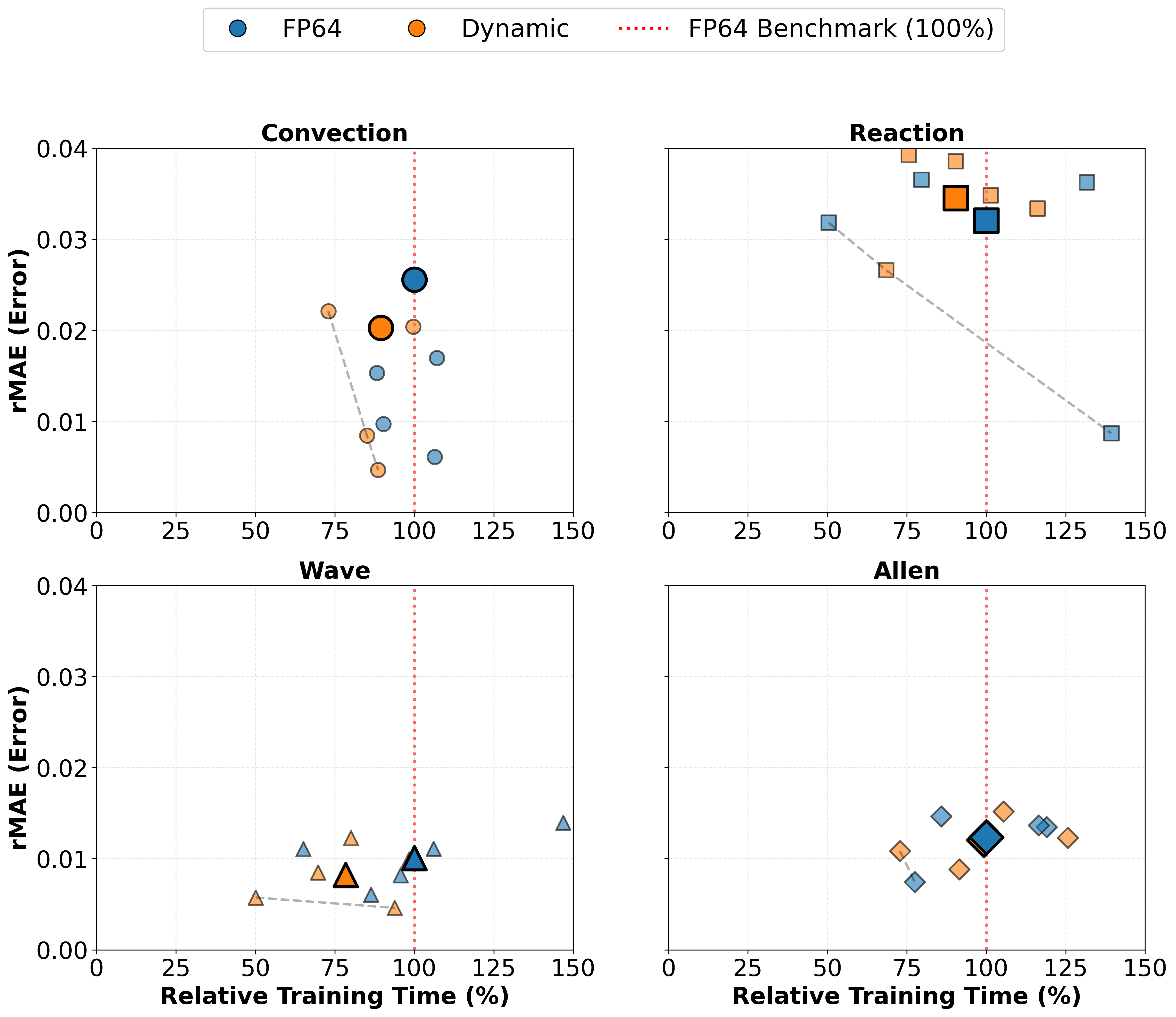}
    \captionsetup{hypcap=false}
    \captionof{figure}{Three-layer Vanilla PINN result plot for dynamic vs.\ FP64. We use 3 hidden layers and 648 hidden dimensions.}
    \label{fig:3-layer_speed_vs_error}
\end{center}

\newpage
\section{PINNsFormer results}
\resetappendixcounters
To evaluate whether the dynamic approach generalises to different neural network architectures, we compare its training time and accuracy using PINNsFormer and PINNMamba. The proposed dynamic approach use lower training time and reach similar level of prediction accuracy.

\begin{center}
\captionsetup{hypcap=false}
\captionof{table}{PINNsFormer result from 3 random seeds.}
\begin{adjustbox}{max width=0.99\textwidth}
\begin{tabular}{cccccc}
\toprule
Equation & approach & Avg Time (s) & rRMSE & rMAE & Speedup vs FP64 \\
\midrule
\textcolor{blue}{\textit{Convection}} & \textcolor{blue}{dynamic} & \textcolor{blue}{\(\mathbf{2,142.51} \pm \mathbf{744.54}\)} & \textcolor{blue}{\(0.0079 \pm 0.0028\)} & \textcolor{blue}{\(0.0069 \pm 0.0026\)} & \textcolor{blue}{\(\mathbf{1.12}\times\)} \\
\textit{Convection} & FP32 & \(936.01 \pm 243.92\) & \(0.2071 \pm 0.3234\) & \(0.1806 \pm 0.2807\) & \(2.57\times\) \\
\textit{Convection} & FP64 & \(2,405.80 \pm 347.89\) & \(\textbf{0.0073} \pm 0.0029\) & \(\textbf{0.0065} \pm 0.0022\) & \(1.00\times\) \\
\hline
\textcolor{blue}{\textit{Reaction}} & \textcolor{blue}{dynamic} & \textcolor{blue}{\(323.94 \pm 106.70\)} & \textcolor{blue}{\(0.0260 \pm 0.0062\)} & \textcolor{blue}{\(0.0130\pm 0.0033\)} & \textcolor{blue}{\(1.33\times\)} \\
\textit{Reaction} & FP32 & \(167.74 \pm 23.31\) & \(0.0320 \pm 0.0024\) & \(0.0160 \pm 0.0014\) & \(\mathbf{2.57}\times\) \\
\textit{Reaction} & FP64 & \(431.00 \pm 103.90\) & \(\textbf{0.0211} \pm \textbf{0.0015}\) & \(\textbf{0.0103} \pm \textbf{0.0007}\) & \(1.00\times\) \\
\hline
\textcolor{blue}{\textit{Wave}} & \textcolor{blue}{dynamic} & \textcolor{blue}{\(\mathbf{34,280.48} \pm \mathbf{3,766.63}\)} & \textcolor{blue}{\(\mathbf{0.0134} \pm \mathbf{0.0051}\)} & \textcolor{blue}{\(\mathbf{0.0127} \pm \mathbf{0.0049}\)} & \textcolor{blue}{\(\mathbf{1.03}\times\)} \\
\textit{Wave} & FP32 & \(3,511.79 \pm 641.10\) & \(0.3429 \pm 0.0244\) & \(0.3345 \pm 0.0286\) & \(10.08\times\) \\
\textit{Wave} & FP64 & \(35,384.31 \pm 4,874.25\) & \(0.0161 \pm 0.0042\) & \(0.0154 \pm 0.0041\) & \(1.00\times\) \\
\hline
\textcolor{blue}{\textit{Allen--Cahn}} & \textcolor{blue}{dynamic} & \textcolor{blue}{\(\mathbf{13,441.23} \pm \mathbf{2,526.00}\)} & \textcolor{blue}{\(\ 0.0150\pm 0.0017\)} & \textcolor{blue}{\(0.0038 \pm 0.0004\)} & \textcolor{blue}{\(\mathbf{1.05}\times\)} \\
\textit{Allen--Cahn}& FP32 & \(2,961.68 \pm 1,125.14\) & \(0.2031 \pm 0.2414\) & \(0.1131 \pm 0.1637\) & \(4.75\times\) \\
\textit{Allen--Cahn}& FP64 & \(14,072.01 \pm 480.71\) & \(\textbf{0.0119} \pm \textbf{0.0032}\) & \(\textbf{0.0031} \pm \textbf{0.0008}\) & \(1.00\times\) \\
\bottomrule
\end{tabular}\label{tab:curvature_aware_PINNsFormer}
\end{adjustbox}
\end{center}
\begin{center}
\captionof{table}{PINNmamba Average result from 5 random seeds. We train on a simple case (\textit{Wave}) and a difficult case (\textit{Convection}) as example.}
\begin{adjustbox}{max width=0.99\textwidth}
\begin{tabular}{cccccc}
\toprule
Equation & approach & Avg Time (s) & rRMSE & rMAE & Speedup vs FP64 \\
\midrule
\textcolor{blue}{\textit{Wave}} & \textcolor{blue}{dynamic} & \textcolor{blue}{\(\mathbf{8,818.70} \pm \mathbf{1,160.00}\)} & \textcolor{blue}{\(\mathbf{0.0063} \pm \mathbf{0.0021}\)} & \textcolor{blue}{\(\mathbf{0.0062} \pm \mathbf{0.0021}\)} & \textcolor{blue}{\(\mathbf{1.28}\times\)} \\
\textit{Wave} & FP32 & \(1,117.12 \pm 732.88\) & \(0.3401 \pm 0.3972\) & \(0.3395 \pm 0.4040\) & \(10.13\times\) \\
\textit{Wave} & FP64 & \(11,311.93 \pm 3,583.74\) & \(0.0077 \pm 0.0050\) & \(0.0077 \pm 0.0050\) & \(1.00\times\) \\
\hline
\textcolor{blue}{\textit{Convection}} & \textcolor{blue}{dynamic} & \textcolor{blue}{\(\mathbf{2,763.09} \pm \mathbf{444.70}\)} & \textcolor{blue}{\(0.0196 \pm 0.0114\)} & \textcolor{blue}{\(0.0172 \pm 0.0103\)} & \textcolor{blue}{\(\mathbf{1.09}\times\)} \\
\textit{Convection} & FP32 & \(168.19 \pm 111.32\) & \(1.3330 \pm 0.3474\) & \(1.2088 \pm 0.3084\) & \(17.87\times\) \\
\textit{Convection} & FP64& \(3,005.28 \pm 613.26\) & \(\textbf{0.0142} \pm \textbf{0.0062}\) & \(\textbf{0.0128} \pm \textbf{0.0051}\) & \(1.00\times\) \\
\bottomrule
\end{tabular}\label{tab:wave_PINNmamba}
\end{adjustbox}
\end{center}

\section{Irradiance ODE prediction} \label{sec:irradiance_result}
\resetappendixcounters
The prediction result comparison for the irradiance ODE prediction is presented in Table~\ref{tab:irradiance_ode}.

\begin{center}
\captionsetup{hypcap=false}
\captionof{table}{Average irradiance result from 5 random seeds.}
\label{tab:irradiance_ode}
\begin{tabular}{cccccc}
\toprule
Equation & approach & Avg Time (s) & rRMSE & rMAE & Speedup vs FP64 \\
\midrule
\multirow{3}{*}{Irradiance ODE (Eq.~\ref{eq:irradiance})}
& \textcolor{blue}{dynamic} 
& \textcolor{blue}{\(78.82 \pm 10.53\)} 
& \textcolor{blue}{\(\textbf{0.0098} \pm \textbf{0.0025}\)} 
& \textcolor{blue}{\(\textbf{0.0083} \pm 0\textbf{.0020}\)} 
& \textcolor{blue}{\(\textbf{1.1}\times\)} \\
& FP32 
& \(75.30 \pm 21.34\) 
& \(0.1079 \pm 0.2142\) 
& \(0.0919 \pm 0.1818\) 
& \(1.15\times\) \\
& FP64 
& \(86.50 \pm 12.93\) 
& \(0.0107 \pm 0.0047\) 
& \(0.0092 \pm 0.0043\) 
& \(1.0\times\) \\
\bottomrule
\end{tabular}
\end{center}

\begin{center}
\begin{minipage}{\linewidth}
\centering
\includegraphics[width=0.98\linewidth]{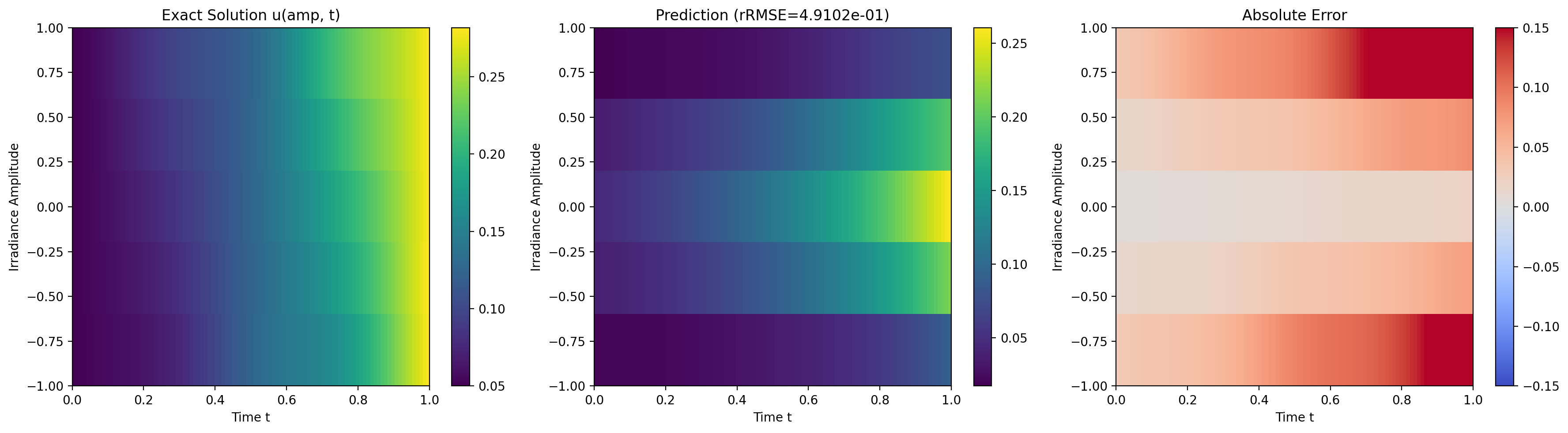}\\[-0.25em]
{\small (a) FP32}
\vspace{0.4em}
\includegraphics[width=0.98\linewidth]{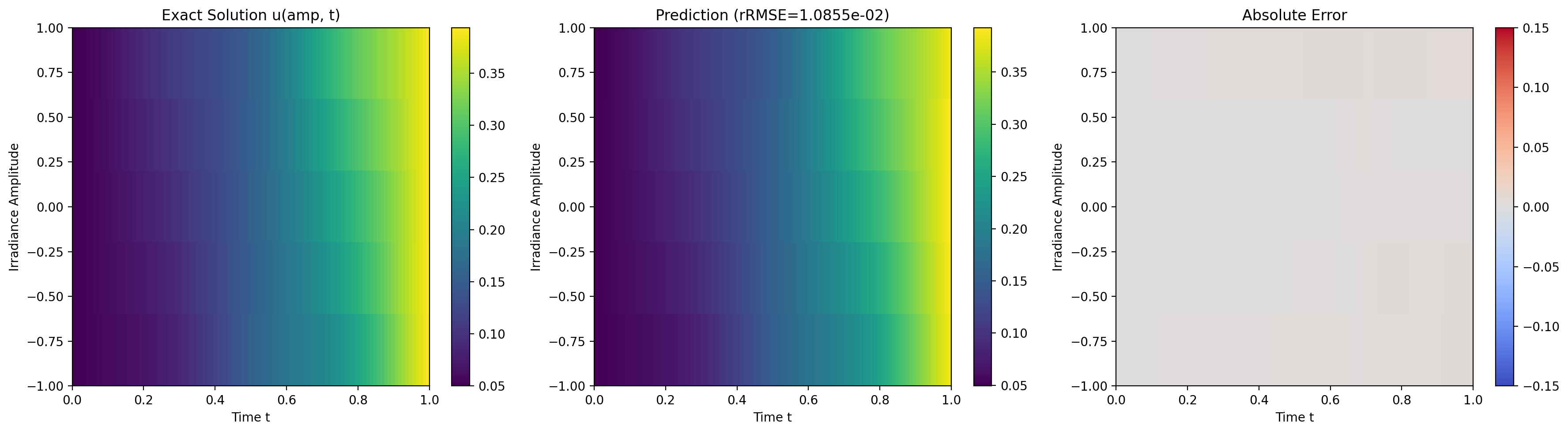}\\[-0.25em]
{\small (b) Dynamic}
\captionsetup{hypcap=false}
\captionof{figure}{Irradiance ODE results: ground truth (left), prediction (middle), and error (right). Single-precision training stops with a relatively large error under a stronger irradiance effect (top), whereas the proposed dynamic-precision approach converges to a smaller error (bottom).}
\label{fig:irradiace_fp32_dynamic}
\end{minipage}
\end{center}
\begin{center}
\begin{minipage}{\linewidth}
\centering

\includegraphics[width=0.31\linewidth]{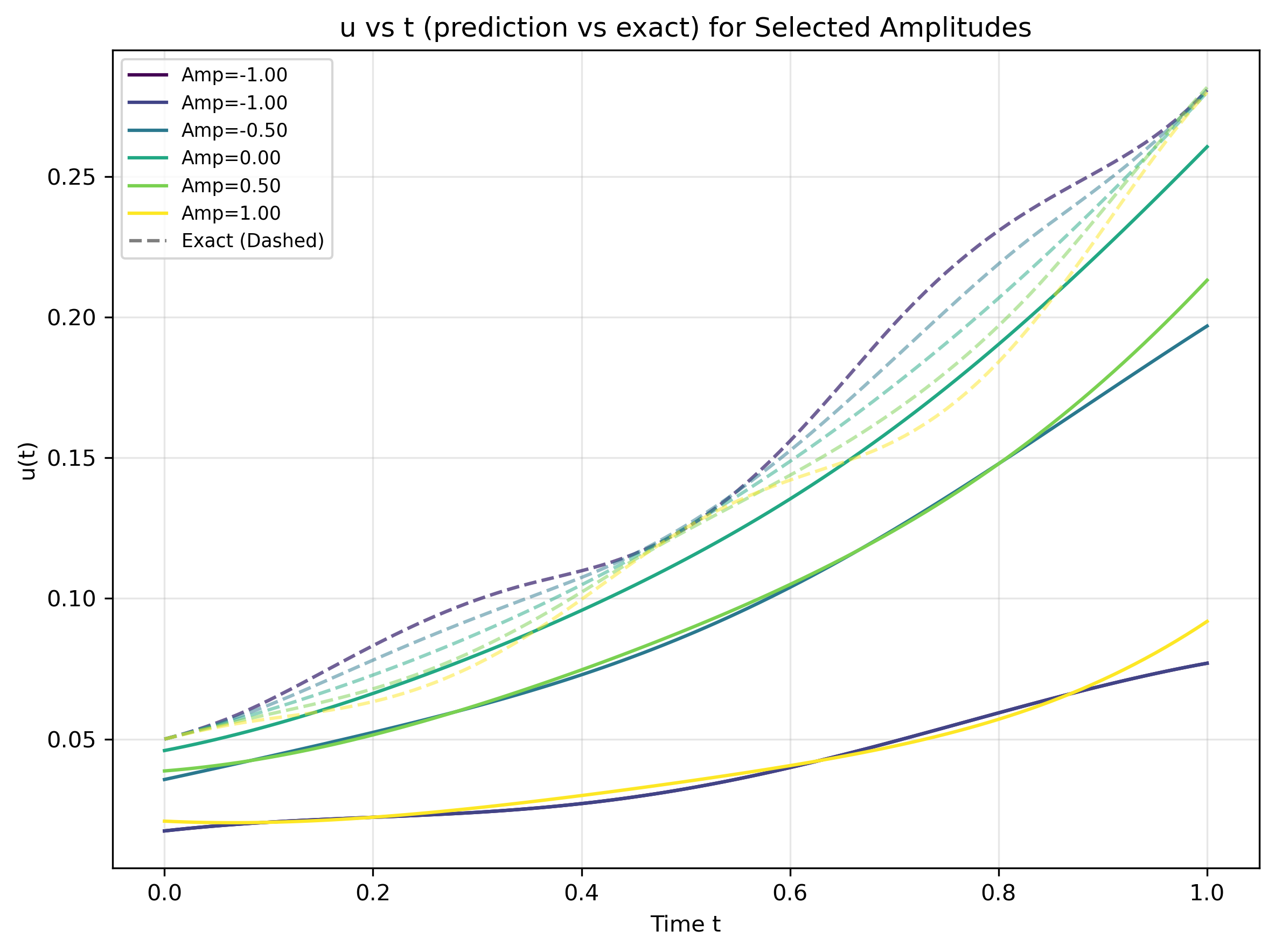}
\hfill
\includegraphics[width=0.31\linewidth]{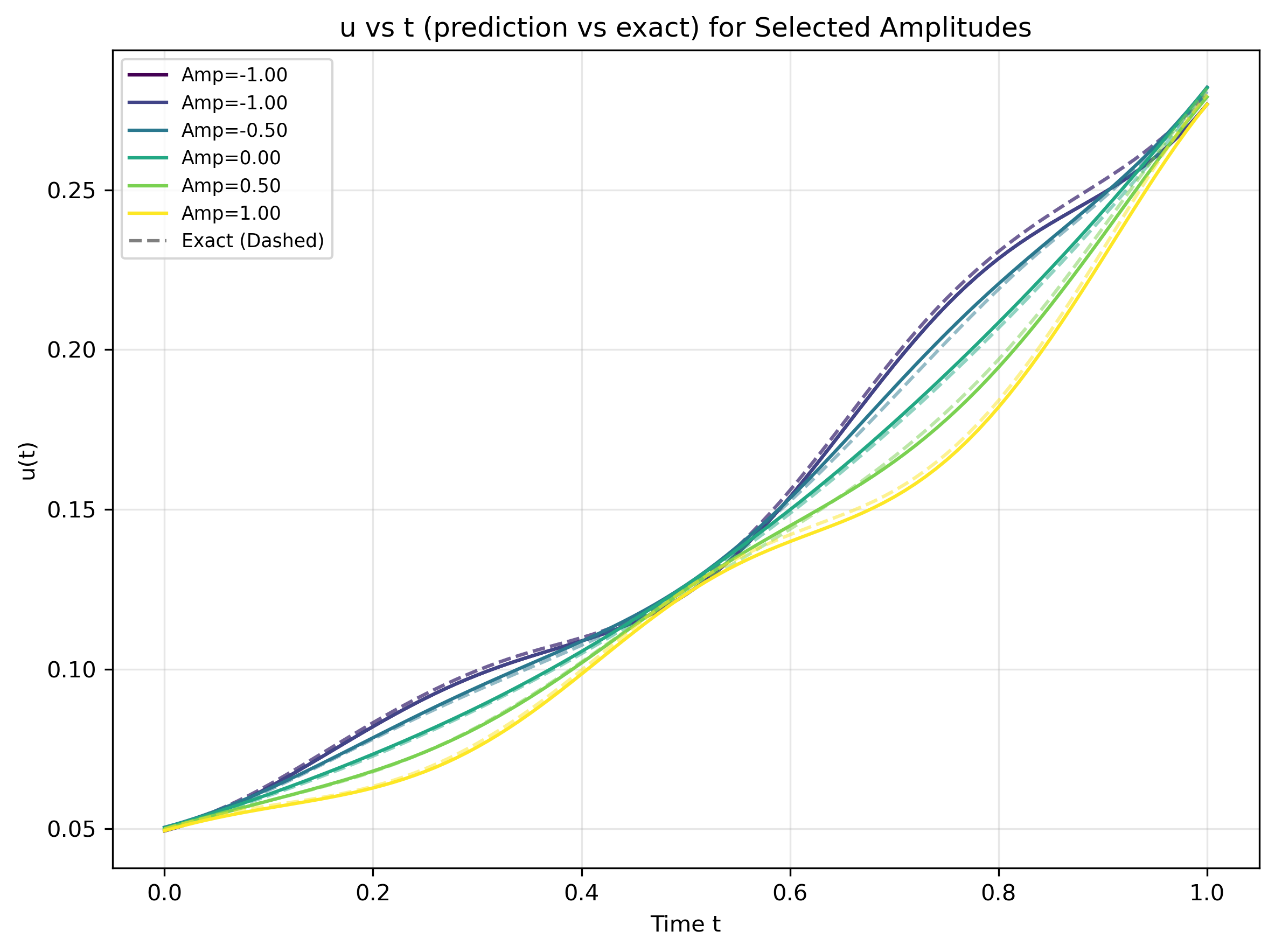}
\hfill
\includegraphics[width=0.31\linewidth]{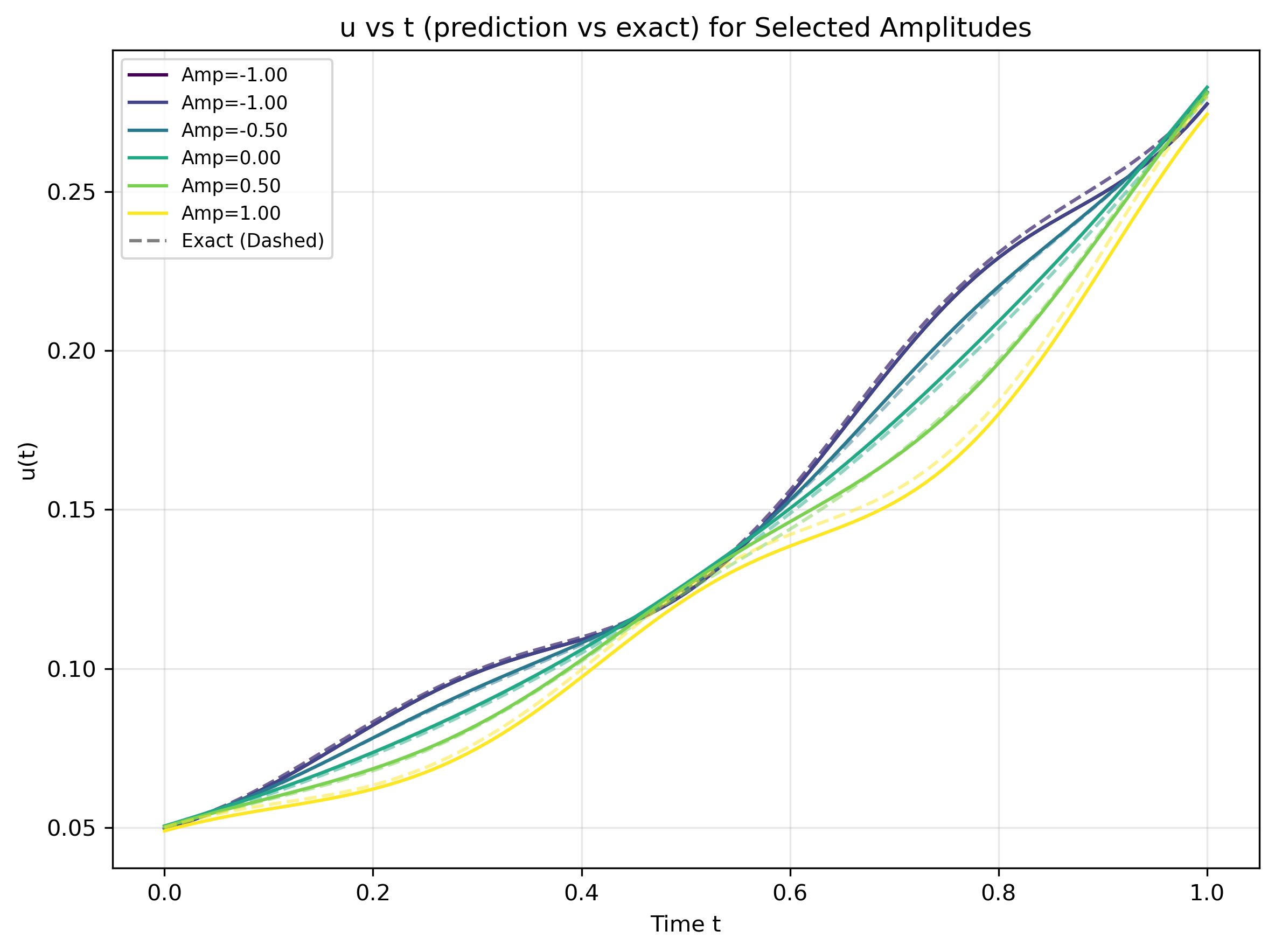}

\vspace{-0.25em}

\begin{minipage}{0.31\linewidth}
\centering
{\small (a) FP32}
\end{minipage}
\hfill
\begin{minipage}{0.31\linewidth}
\centering
{\small (b) FP64}
\end{minipage}
\hfill
\begin{minipage}{0.31\linewidth}
\centering
{\small (c) Dynamic}
\end{minipage}

\captionsetup{hypcap=false}
\captionof{figure}{Irradiance ODE results comparison: FP32, FP64, and dynamic precision predictions with the ground truth, shown from left to right. The three examples were run with the same random seed, and FP32 failed to converge to the ground truth.}
\label{fig:irradiance_fp32_fp64_dynamic}

\end{minipage}
\end{center}
\FloatBarrier
\end{appendices}
\newpage
\bibliographystyle{cas-model2-names}
\bibliography{reference_PINN}
\end{document}